\documentclass[11pt]{article}

\usepackage[final]{acl}

\usepackage{url}
\usepackage{times}
\usepackage{latexsym}
\usepackage{algorithm}
\usepackage{algpseudocode}
\usepackage{amsmath}
\usepackage{amssymb}
\usepackage[T1]{fontenc}
\usepackage[utf8]{inputenc}
\usepackage{microtype}
\usepackage{inconsolata}
\usepackage{booktabs}
\usepackage{graphicx}
\usepackage[table]{xcolor}

\definecolor{oursrowgray}{HTML}{EAF1FB}
\definecolor{rankfirst}{HTML}{AFC7E8}
\definecolor{ranksecond}{HTML}{D5E3F5}
\definecolor{rankthird}{HTML}{EDF3FA}
\definecolor{groupgray}{HTML}{E6EBF2}
\newcommand{\firstscore}[1]{\cellcolor{rankfirst}\textbf{#1}}
\newcommand{\secondscore}[1]{\cellcolor{ranksecond}#1}
\newcommand{\thirdscore}[1]{\cellcolor{rankthird}#1}

\title{ResMerge: Residual-based Spectral Merging of Large Language Models}

\author{
\textbf{Yandu Sun}\textsuperscript{1}\thanks{Equal contribution.}\!,
\textbf{Zhiyan Hou}\textsuperscript{2,3}\footnotemark[1]\!,
\textbf{Hongyan An}\textsuperscript{4}\!,
\textbf{Weizhen Wang}\textsuperscript{5}\!,
\textbf{Haokai Ma}\textsuperscript{6}\thanks{Corresponding authors.}\!,\\%
\textbf{Yuheng Jia}\textsuperscript{1}\footnotemark[2]\!,
\textbf{Junfeng Fang}\textsuperscript{6},
\textbf{Haiyun Guo}\textsuperscript{2,3},
\textbf{Jinqiao Wang}\textsuperscript{2,3,7}\\[0.4em]%
\textsuperscript{1}Southeast University, Nanjing, China\\%
\textsuperscript{2}Institute of Automation, Chinese Academy of Sciences\quad
\textsuperscript{3}University of Chinese Academy of Sciences\\%
\textsuperscript{4}Wuhan University of Technology, Wuhan, China\quad
\textsuperscript{5}Peking University, Beijing, China\\%
\textsuperscript{6}National University of Singapore\quad
\textsuperscript{7}Wuhan AI Research, Wuhan, China\\[0.4em]%
{\small
\begin{minipage}{0.95\linewidth}\centering
sunyd0303@gmail.com\quad
houzhiyan23@mails.ucas.ac.cn\quad
haokai.ma@nus.edu.sg\quad
yhjia@seu.edu.cn
\end{minipage}}%
}

\begin{document}
\maketitle

\begin{abstract}
Model merging offers a training-free way to combine multiple post-trained expert models, but merging experts obtained through reinforcement learning (RL) remains challenging. Existing spectral merging methods often assume that leading singular directions contain the main task signal, while lower-energy residual components can be compressed, selected, or attenuated to reduce interference. We find that this assumption does not hold for RL task vectors: after decomposing each task vector into a leading spectral head and a residual component, both parts can independently recover substantial behavior knowledge, while exhibiting different merging properties. The head is highly concentrated and informative but more prone to sharp cross-expert conflicts, whereas the residual component is more dispersed and provides a more stable basis for aggregation. Based on this observation, we propose \textsc{ResMerge}, a residual-based spectral merging framework for RL experts. \textsc{ResMerge} first constructs a stable residual backbone with Spherical Residual Consensus Adaptation, which estimates a reliability-weighted consensus direction on the Frobenius sphere. It then reintroduces leading-head information through a Lightweight Head Correction module gated by positive cross-expert agreement. Experiments across multiple RL expert groups and capability domains show that \textsc{ResMerge} better preserves expert capabilities than representative task-vector and spectral merging baselines.
\end{abstract}

\section{Introduction}

Post-training has become the dominant paradigm for adapting large pretrained models to downstream tasks, from RLHF-style alignment \citep{ouyang2022training} to reasoning-oriented reinforcement learning \citep{guo2025deepseek}, producing many task-specialized expert models. Re-running post-training on the union of their data is costly and hard to scale. Model merging offers a training-free alternative: it combines several experts into one model, from weight averaging such as model soups \citep{wortsman2022model} to task-vector composition such as Task Arithmetic \citep{ilharco2023editing}.

\begin{figure}[t]
\centering
\includegraphics[
width=\linewidth
]{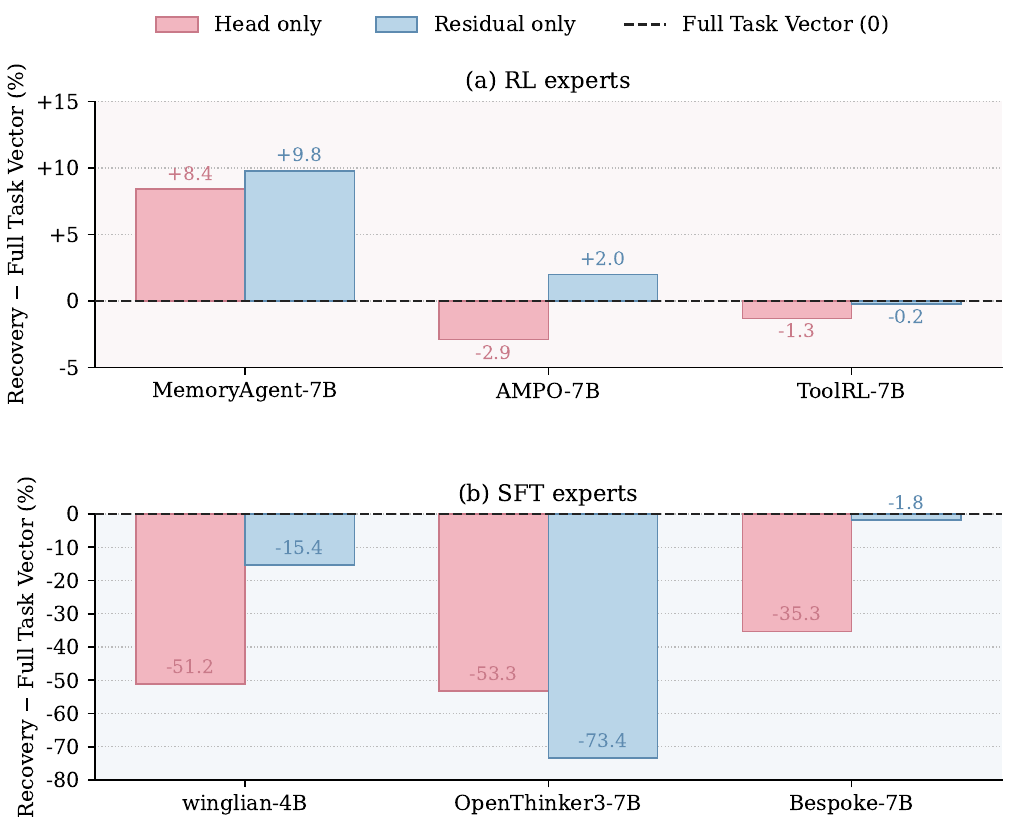}
\caption{
\textbf{Component-level recovery under RL and SFT post-training.}
After applying singular value decomposition (SVD) to each task vector, \emph{Head-only} retains the rank-1 head formed by the top singular direction, while \emph{Residual-only} retains the remaining spectral residual after removing this head.
Both components are more recoverable in RL task vectors than in SFT task vectors, supporting component-wise treatment of RL spectral updates.
}
\label{fig:motivation}
\end{figure}

A central difficulty of model merging is that expert-specific task vectors, defined as the parameter differences between fine-tuned experts and the shared base model, can be poorly aligned in the shared weight-update space \citep{ilharco2023editing}. 
Directly adding or averaging these task vectors can cause negatively aligned components to cancel out, weakening the task-specific abilities each expert was supposed to contribute \citep{yadav2023ties}. 
To reduce such interference, spectral methods apply singular value decomposition (SVD) to task vectors and operate on their singular components. 
TSV-Merge uses task singular vectors to reduce interference \citep{gargiulo2024task}, STAR truncates and rescales spectral components \citep{lee2025star}, AdaRank selects beneficial singular ranks \citep{lee2025adarank}, and SVC calibrates shared spectral directions \citep{li2026shared}. 
These methods generally assume that high-energy leading singular directions carry the main task ability, while the remaining components can be compressed, selected, reweighted, or calibrated to reduce interference.

In our experiments, we find that this spectral assumption does not hold
consistently for reinforcement learning (RL) experts. In particular,
the leading singular directions and the remaining residual component
cannot be cleanly separated into dominant signal and redundant noise.

As shown in Figure~\ref{fig:motivation}, keeping only the leading rank-1 head or only the residual component can both recover much of the full task-vector ability, and sometimes even surpass it. 
This suggests that both components encode behavior knowledge, so the residual component should not be treated as subordinate noise to compress or attenuate.

Building on this observation, we propose \textsc{ResMerge}, a residual-based spectral merging framework for RL experts. It explicitly splits each task vector into a leading spectral head and a residual component, and designs separate merging rules for the two components. Our visualization results show that the head component concentrates energy into a small number of task-specific directions, which makes it informative on its own but also more prone to sharp conflicts across experts. The residual component, by contrast, has a flatter and more dispersed spectrum and behaves more stably under aggregation, making it a better basis for a unified merge. We therefore first use the residual singular directions to construct the merging backbone, and then, using this backbone as a reference, selectively inject the components of the leading head that are compatible with the aggregated update. This way, the task ability carried by the high-energy spectral directions is preserved without letting their unreliable parts dominate the merged model.

Our contributions are threefold: (I) We expose a property of RL task vectors that distinguishes them from SFT task vectors, where the leading spectral head and the residual component can each independently recover, or even surpass, the task ability of the full task vector, indicating that the residual component does not naturally occupy a subordinate position that can be safely compressed or attenuated. (II) We propose a new merging paradigm for RL experts that decomposes each task vector into a leading spectral head and a residual component, constructs the merging backbone from the more stable residual component, and selectively absorbs the part of the leading head that is compatible with this backbone. (III) Experiments across multiple RL expert groups show that our method preserves expert capabilities substantially better than existing merging baselines after combining multiple RL experts into a single model.

\section{Related Work}
\label{sec:related_work}

\paragraph{Model merging.}
Model merging aims to combine multiple expert models fine-tuned for different tasks into a single model with multi-task capabilities, without additional training or inference-time overhead. One major challenge in model merging is task interference, where updates from different experts may conflict, cancel useful knowledge, or over-amplify shared directions. Early weight-space methods show that models fine-tuned from the same initialization can often be averaged effectively \citep{wortsman2022model}. Task Arithmetic formalizes this idea by defining a task vector $\Delta_i = W_i - W_0$ and composing model behaviors through linear operations over such vectors \citep{ilharco2023editing}. Methods such as TIES-Merging mitigate this issue by resolving parameter-level redundancy and sign conflicts \citep{yadav2023ties}. 
\paragraph{SVD for model merging.}
More recent spectral approaches analyze task matrices through SVD. TSV-Merge uses task singular vectors to reduce interference \citep{gargiulo2024task}; STAR truncates and rescales spectral components \citep{lee2025star}; AdaRank adaptively selects beneficial singular ranks \citep{lee2025adarank}; and SVC calibrates inflated singular values caused by shared spectral directions \citep{li2026shared}. These methods highlight the importance of spectral structure, but largely treat singular components as homogeneous units to select, compress, or calibrate. We instead show that RL task vectors exhibit functional heterogeneity across spectral components: leading singular directions and residual singular directions both encode recoverable behavior knowledge, yet require different merging rules due to their differences in energy concentration, geometric consistency, and merging risk.

\paragraph{Merging RL post-trained models.}
Recent work has started to study model merging in RL or RLHF post-training scenarios. 
Rewarded Soups interpolates reward-tuned policies to improve alignment trade-offs \citep{rame2023rewarded}, while WARP uses weight-space operations such as EMA, spherical interpolation, and interpolation toward initialization to improve the reward--KL trade-off in RLHF \citep{rame2024warp}. 
SALSA studies soup-based alignment learning for stronger RLHF adaptation \citep{huang2024salsa}, and LiNeS uses layer-wise scaling to reduce forgetting and interference in model merging, including the merging of RLHF-aligned language models \cite{wang2025lines}. 
These methods show that RL- or preference-trained models can benefit from weight-space merging, but they mainly focus on alignment objectives and policy-level trade-offs, rather than examining the internal structure of RL task vectors in detail.
RAM is the most related work: it studies merging RL-trained agentic models and decomposes updates into shared and unique behavior knowledge based on parameter-level patterns \citep{yuan2026behavior}. 
In contrast, we focus on spectral heterogeneity within each RL task vector, where leading singular directions and residual singular directions both retain behavior knowledge but exhibit different stability and merging risks.

\section{Preliminaries}
\label{sec:preliminaries}

\subsection{Model Merging}
\label{sec:prelim_model_merging}

We consider $N$ expert models $\{W_i\}_{i=1}^{N}$ sharing the same base model $W_0$ and specialized for different tasks, domains, or behavioral objectives. Model merging aims to combine these experts into a single model $W_{\mathrm{merge}}$ without additional training or inference-time ensembling \citep{wortsman2022model, ilharco2023editing}:
\begin{equation}
W_{\mathrm{merge}}
= \mathcal{M}(W_0, \{W_i\}_{i=1}^{N}),
\end{equation}
where $\mathcal{M}(\cdot)$ denotes a specific merging rule, and $W_{\mathrm{merge}}$ is expected to effectively retain the key capabilities of all experts.

\subsection{Task Arithmetic}
\label{sec:prelim_task_arithmetic}

Task arithmetic is a widely used model merging paradigm that composes model behaviors through arithmetic operations on task vectors \citep{ilharco2023editing}. Given an expert model $W_i$ and the shared base model $W_0$, the corresponding task vector is defined as the parameter difference
\begin{equation}
\Delta_i = W_i - W_0.
\end{equation}
For layer-wise merging, we further write
\begin{equation}
\Delta_i^{(l)} = W_i^{(l)} - W_0^{(l)},
\end{equation}
where $\Delta_i^{(l)}$ is the task matrix at layer $l$. We omit the layer superscript when clear. A standard task-arithmetic merge is
\begin{equation}
W_{\mathrm{TA}}
= W_0 + \lambda \sum_{i=1}^{N} \Delta_i,
\label{ta}
\end{equation}
where $\lambda$ is a scaling coefficient. Task arithmetic is simple yet often effective, and is commonly used as a baseline in model merging.

\subsection{Singular Value Decomposition}
\label{sec:prelim_svd}

Singular value decomposition (SVD) is a standard matrix factorization technique that represents a matrix as a sum of rank-one components ordered by their corresponding singular values. For a task matrix $\Delta_i \in \mathbb{R}^{m \times n}$, SVD gives the following.
\begin{equation}
\Delta_i
= U_i \Sigma_i V_i^\top
= \sum_{r=1}^{R}
\sigma_{i,r} u_{i,r} v_{i,r}^\top,
\end{equation}
where $U_i$ and $V_i$ contain the left and right singular vectors, $\Sigma_i$ is a diagonal matrix of singular values, $\sigma_{i,r}$ is the $r$-th singular value with $\sigma_{i,1}\geq\sigma_{i,2}\geq\cdots$, $u_{i,r}$ and $v_{i,r}$ are the corresponding singular vectors, and $R$ is the matrix rank. Each rank-one term $\sigma_{i,r}u_{i,r}{v_{i,r}}^\top$ corresponds to one spectral direction of the task matrix. Components associated with larger singular values form the leading singular directions, while the remaining components form residual singular directions.

\begin{figure*}[!t]
\centering
\includegraphics[
  width=0.95\textwidth
]{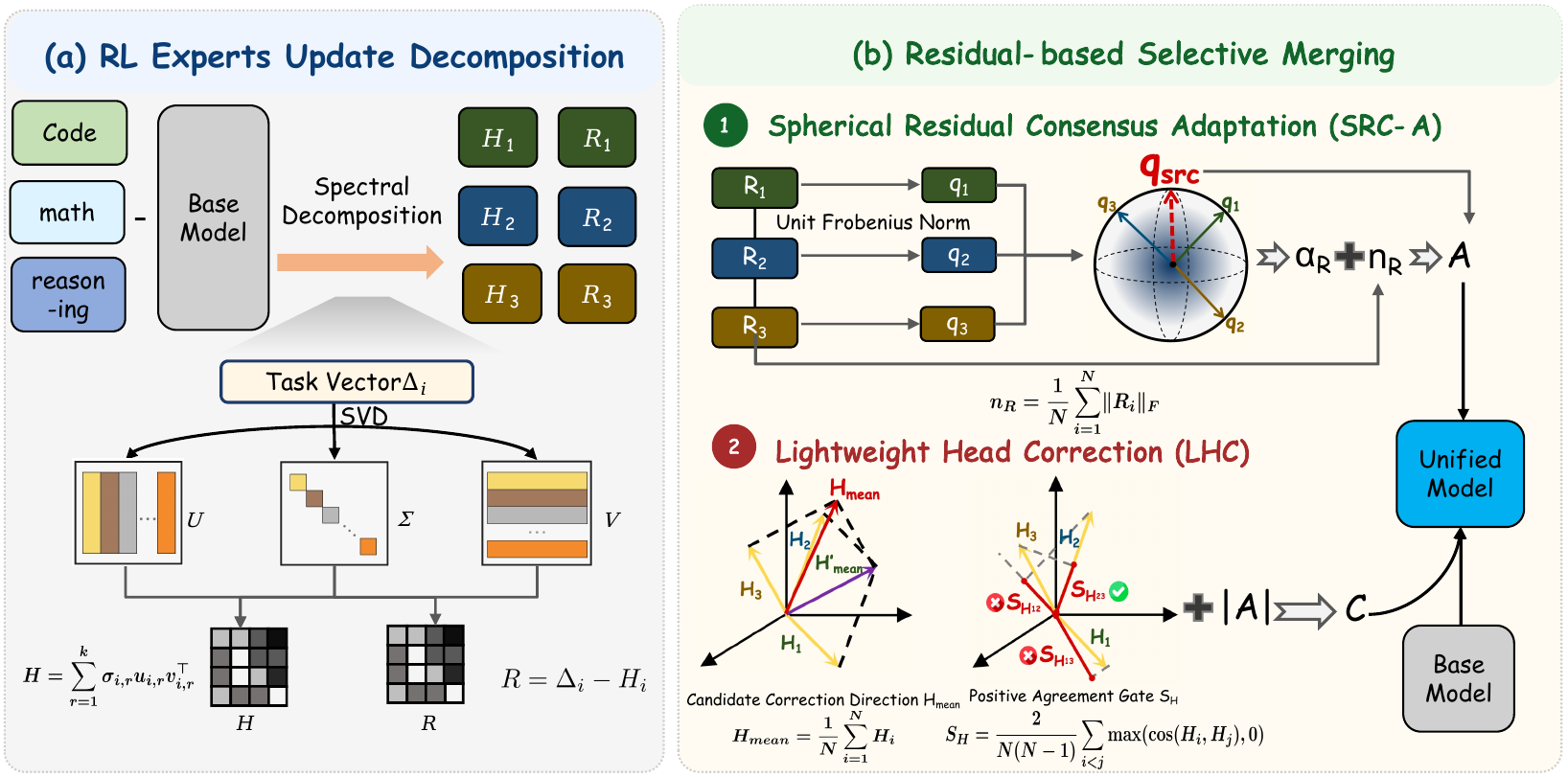}
\caption{
\textbf{Overview of \textsc{ResMerge}. }
Given RL expert task vectors, we decompose each update into a rank-1 spectral head and a residual component. 
The residual components are merged into a stable SRC-A backbone, while the rank-1 heads are added back only as reliability-gated lightweight corrections.
}
\label{fig:method_overview}
\end{figure*}

\section{Method}
\label{sec:method}

In this section, we propose a residual-based spectral merging method for RL models, which builds a robust and stable backbone from residual singular directions with Spherical Residual Consensus Adaptation (SRC-A) and selectively injects informative leading singular directions through Lightweight Head Correction (LHC) to improve specialization.

\subsection{Layer-wise Spectral Decomposition}
\label{sec:layer_spectral_decomposition}

For each mergeable matrix-shaped layer, we apply SVD to the task matrix $\Delta_i$ using the notation introduced in Section~\ref{sec:prelim_svd}, and omit the layer index for readability. We then split $\Delta_i$ into a head $H_{i,k}$ containing the top-$k$ leading singular directions and a residual component $R_{i,k}$ containing the remaining lower-energy residual singular directions:
\begin{equation}
H_{i,k}
= \sum_{r=1}^{k}
\sigma_{i,r} u_{i,r} v_{i,r}^{\top},
\quad
R_{i,k}
= \Delta_i - H_{i,k}.
\end{equation}
Here, $k$ is the number of leading singular directions retained in the spectral head component; we omit $k$ when the context is clear.

\subsection{Spherical Residual Consensus Adaptation}
\label{sec:srca}
After removing the leading singular directions, the residual component contains lower-energy and more dispersed directions, making its merging risk less concentrated than the head. We therefore use it as the backbone, but avoid plain Euclidean averaging because it may dilute weak shared directions. SRC-A uses the average residual norm as a scale anchor and estimates a consensus direction on the Frobenius unit sphere.

We use the Frobenius cosine $\cos(X,Y)=\langle X,Y\rangle_F/(\|X\|_F\|Y\|_F+\epsilon)$ to measure directional agreement between matrix-shaped updates, where $\epsilon$ is a small constant for numerical stability. Each residual component is normalized as
\begin{equation}
q_i = \frac{R_i}{\|R_i\|_F + \epsilon}
\;,\;
q_{\mathrm{src}} =
\frac{\sum_{i=1}^{N} q_i}
{\left\|\!\sum_{i=1}^{N} q_i\!\right\|_F + \epsilon}.
\end{equation}
Here, $q_i$ keeps only the normalized direction of the residual component $R_i$, and $q_{\mathrm{src}}$ is the resulting spherical consensus direction.

We estimate the reliability of this consensus using pairwise residual agreement $s_R$ and consensus alignment $c_R$:
\begin{equation}
\begin{aligned}
s_R &=
\frac{2}{N(N-1)}
\sum_{i<j} \cos(q_i, q_j), \\
c_R &=
\frac{1}{N}\sum_{i=1}^{N}
\cos(q_i, q_{\mathrm{src}}).
\end{aligned}
\end{equation}
They measure mutual residual alignment and consensus agreement,
respectively. For computing the overall reliability score, we map both
quantities to $\bar{s}_R,\bar{c}_R\in[0,1]$ using $(x+1)/2$ and define
$\mathrm{score}_R=(\bar{s}_R+\bar{c}_R)/2$.
The resulting score determines a bounded retention exponent:
\begin{equation}
\beta = \beta_{\max} - (\beta_{\max}-\beta_{\min})\mathrm{score}_R,
\end{equation}
where $\beta_{\min}$ and $\beta_{\max}$ keep the retention exponent within a conservative range. We use the average residual norm to provide a stable scale reference:
\begin{equation}
n_R =
\frac{1}{N}
\sum_{i=1}^{N}
\|R_i\|_F.
\end{equation}

We use the original consensus alignment $c_R$, rather than its mapped
version $\bar{c}_R$, to control the retained residual magnitude.
Specifically, we define the residual retention coefficient
$\alpha_R$ as
\begin{equation}
\alpha_R = ({c}_R)^{\beta}.
\end{equation}

The residual backbone is then constructed as
\begin{equation}
A = \mathrm{SRC\text{-}A}(\{R_i\}_{i=1}^{N})
= \alpha_R n_R q_{\mathrm{src}}.
\end{equation}

This separates scale, direction, and reliability: $n_R$ anchors the residual scale, $q_{\mathrm{src}}$ gives the consensus direction, and $\alpha_R$ controls the retained residual magnitude. In particular, a smaller $c_R$ directly reduces
$\alpha_R$, suppressing the residual update under weak consensus,
whereas stronger consensus preserves more magnitude along
$q_{\mathrm{src}}$.

\subsection{Lightweight Head Correction}
The head formed by leading singular directions can encode important behavior knowledge, but its highly concentrated structure makes merging risk more localized and severe. Since poorly aligned heads may yield an unreliable shared direction under direct averaging, we use the head only as a lightweight correction gated by positive cross-expert agreement among different heads.

We first compute the average head as a candidate correction direction:
\begin{equation}
H_{\mathrm{mean}}
= \frac{1}{N}\sum_{i=1}^{N} H_i.
\end{equation}
To assess whether this direction is reliable across experts, we measure positive agreement among individual heads:
\begin{equation}
S_H =
{\textstyle \frac{2}{N(N-1)}
\sum_{i<j}
\max(\cos(H_i,H_j),0)} .
\end{equation}

The raw norm of the head reflects how strongly an expert moves along its leading singular directions, but it does not indicate whether this movement is reliable for multi-expert merging. A high-norm head may still be task-specific or poorly aligned with other experts. Directly adding the averaged head according to its own norm would therefore couple the correction strength to expert-specific magnitude rather than cross-expert agreement. To avoid this, we decouple direction, reliability, and scale:
\begin{equation}
C =
\rho \, S_H \, \|A\|_F
\frac{H_{\mathrm{mean}}}{\|H_{\mathrm{mean}}\|_F+\epsilon}.
\end{equation}
Here, $\rho$ is the maximum injection ratio. The correction is gated by head agreement through $S_H$ and scaled by the residual backbone norm $\|A\|_F$. Thus, consistent heads provide a controlled behavior-specific correction, while poorly aligned heads are suppressed as $S_H$ approaches zero.

\subsection{Final Merging Rule}
For each mergeable matrix-shaped layer, the final merged update is the sum
of the SRC-A residual backbone and the lightweight head correction term.
The merged parameters are then obtained by adding this update to the base
model:
\begin{equation}
W_{\mathrm{merge}}^{(l)}
= W_0^{(l)} + A^{(l)} + C^{(l)}.
\end{equation}
For embeddings, normalization parameters, and one-dimensional vectors,
we instead use simple parameter averaging across the $N$ experts:
\begin{equation}
W_{\mathrm{merge}}
=
\frac{1}{N}\sum_{i=1}^{N} W_i.
\end{equation}
This is equivalent to the task-arithmetic update in Eq.~\ref{ta} with
$\lambda=1/N$.

\begin{table*}[t]
\centering
\scriptsize
\renewcommand{\arraystretch}{1.25}
\setlength{\tabcolsep}{3pt}
\resizebox{\textwidth}{!}{
\begin{tabular}{lccccccccccccc}
\toprule
\multicolumn{1}{c}{Method}
& \multicolumn{4}{c}{Tool Using}
& \multicolumn{4}{c}{Math}
& Reasoning
& \multicolumn{3}{c}{Coding}
& Overall \\
\cmidrule(lr){2-5}
\cmidrule(lr){6-9}
\cmidrule(lr){10-10}
\cmidrule(lr){11-13}
\cmidrule(lr){14-14}
&
Live Para & Live P-Mul & Non-Live Para & Non-Live P-Mul
& AIME24 & AIME25 & AMC23 & MATH500
& GPQA-D
& LiveCodeBench & HumanEval++ & MBPP++
& Overall Avg. \\
\midrule
\rowcolor{groupgray}
\multicolumn{14}{c}{\textit{Base and Expert Models}} \\
\textbf{Qwen2.5-7B} & 56.25 & 29.17 & 63.00 & 34.00 & 6.67 & 0.00 & 32.50 & 54.80 & 30.13 & 8.22 & 70.73 & 35.19 & 34.32 \\
\textbf{SimpleRL} & 56.25 & 20.83 & 47.00 & 16.00 & 20.00 & 10.00 & 55.00 & 73.20 & 30.81 & 11.35 & 71.34 & 64.02 & 38.57 \\
\textbf{Zero} & 6.25 & 0.00 & 1.50 & 1.50 & 16.67 & 10.00 & 55.00 & 72.00 & 30.30 & 15.07 & 57.32 & 61.90 & 28.95 \\
\textbf{Reasoner} & 56.25 & 29.17 & 56.50 & 22.00 & 20.00 & 10.00 & 55.00 & 60.80 & 35.35 & 16.63 & 74.39 & 63.23 & 41.05 \\
\midrule
\rowcolor{groupgray}
\multicolumn{14}{c}{\textit{Merged Models}} \\
\textbf{TA} & \secondscore{50.00} & \secondscore{25.00} & 37.50 & \thirdscore{17.50} & \firstscore{20.00} & 3.33 & \secondscore{60.00} & 73.20 & 33.67 & \secondscore{14.29} & 70.73 & 62.96 & 38.66 \\
\textbf{TIES} & \thirdscore{0.00} & 4.17 & 0.00 & 0.50 & \firstscore{20.00} & \firstscore{20.00} & 55.00 & 72.60 & \firstscore{38.72} & 13.11 & 70.12 & \thirdscore{65.34} & 32.83 \\
\textbf{DARE + TIES} & \secondscore{50.00} & \thirdscore{8.33} & 4.00 & 2.00 & \thirdscore{13.33} & \thirdscore{13.33} & \thirdscore{57.50} & \thirdscore{73.80} & \secondscore{38.22} & \thirdscore{13.70} & 69.51 & \firstscore{67.20} & 35.98 \\
\textbf{TSV-Merge} & \secondscore{50.00} & \thirdscore{8.33} & 10.50 & 2.00 & \firstscore{20.00} & 6.67 & \secondscore{60.00} & \secondscore{74.00} & 36.53 & 12.33 & \thirdscore{71.95} & \secondscore{66.14} & 36.14 \\
\textbf{ISO-C} & \firstscore{56.25} & \secondscore{25.00} & \firstscore{44.50} & \firstscore{24.00} & \secondscore{16.67} & 10.00 & 47.50 & 72.20 & 30.47 & \secondscore{14.29} & \firstscore{73.17} & \thirdscore{65.34} & \secondscore{38.86} \\
\textbf{ISO-CTS} & \firstscore{56.25} & \secondscore{25.00} & \thirdscore{43.00} & \secondscore{21.50} & \thirdscore{13.33} & 10.00 & 52.50 & 72.40 & 31.48 & \thirdscore{13.70} & \secondscore{72.56} & 64.02 & \thirdscore{38.77} \\
\textbf{RAM} & \thirdscore{0.00} & 4.17 & 0.00 & 0.50 & \secondscore{16.67} & \firstscore{20.00} & \firstscore{62.50} & 73.60 & \thirdscore{37.21} & 13.50 & 71.34 & 62.43 & 32.67 \\
\midrule
\textbf{\textsc{ResMerge} (Ours)} & \firstscore{56.25} & \firstscore{29.17} & \secondscore{43.50} & \secondscore{21.50} & \secondscore{16.67} & \secondscore{16.67} & \thirdscore{57.50} & \firstscore{74.20} & 34.34 & \firstscore{16.63} & 71.34 & \thirdscore{65.34} & \firstscore{41.08} \\
\bottomrule
\end{tabular}
}
\caption{Main results on the Qwen2.5-7B-Base expert group across individual benchmarks and the overall average. The upper block reports the base model and single-task experts, while the lower block compares merging baselines with our \textsc{ResMerge} method (SRC-A+LHC). Shaded cells denote the top distinct results among merged models in each column: \colorbox{rankfirst}{\textbf{1st}}, \colorbox{ranksecond}{2nd}, and \colorbox{rankthird}{3rd}; tied scores share the same color.}
\label{tab:main_results}
\end{table*}

\section{Experiments}
\label{sec:experiments}

\subsection{Experimental Setup}
\label{sec:exp_setup}

\paragraph{Models and Expert Sets.}
We merge only models that share the same architecture and base initialization. The main table reports the Qwen2.5-7B-Base expert group, consisting of Qwen2.5-7B-Base \citep{yang2024qwen25}, the math expert Qwen-2.5-7B-SimpleRL-Zoo (SimpleRL) \citep{zeng2025simplerl}, the math-reasoning expert Open-Reasoner-Zero-7B (Zero) \citep{hu2025openreasonerzero}, and the general-reasoning expert General-Reasoner-Qwen2.5-7B (Reasoner) \citep{ma2025generalreasoner}. Additional expert groups and model-task mappings are provided in Appendix~\ref{sec:app_setup}.

\paragraph{Datasets.}
We evaluate model performance across five major capability domains in total: coding, general reasoning, mathematics, tool use, and memory. Coding is evaluated with \textsc{LiveCodeBench} release\_v2 \citep{jain2024livecodebench}, \textsc{HumanEvalPlus}, and \textsc{MBPPPlus} \citep{chen2021evaluating, austin2021program, liu2023evalplus}; general reasoning with \textsc{GPQA-Diamond} \citep{rein2023gpqa}; mathematics with \textsc{AIME24}, \textsc{AIME25}, \textsc{AMC23}, and \textsc{MATH500} \citep{hendrycks2021measuring, lightman2023lets}; tool use with \textsc{BFCL V3} \citep{pmlr-v267-patil25a}; and memory with \textsc{HotpotQA} and \textsc{SQuAD} for the instruction-tuned model group \citep{yang2018hotpotqa, rajpurkar2016squad}.

\paragraph{Baselines.}
We compare with task-vector methods, including TA \citep{ilharco2023editing}, TIES-Merging \citep{yadav2023ties}, and DARE \citep{yu2023language}; SVD-based methods, including TSV-Merge \citep{gargiulo2024task}, ISO-C, and ISO-CTS \citep{marczak2025notask}; and the RL expert merging method RAM \citep{yuan2026behavior}.

\paragraph{Implementation Details.}
For our method, SVD is applied to mergeable
matrix-shaped layers, while embeddings, normalization parameters,
and one-dimensional vectors are merged by simple parameter averaging. Unless otherwise specified, we use $k=1$, $\beta_{\min}=0.7$, $\beta_{\max}=1.0$, and $\rho=0.20$ across layers. Our implementation is publicly available at 
\url{https://github.com/sunyd0303-cpu/ResMerge-release}.

\paragraph{Metrics.}
We report category averages and the overall average across domains. We use Pass@1 for coding, accuracy for mathematical reasoning, general reasoning, and memory-oriented QA, and official BFCL metrics for tool use. Full model sources, dataset details, baseline hyperparameters, and metric aggregation rules are provided in Appendix~\ref{sec:app_setup} and Appendix~\ref{sec:app_impl_details}.

\begin{table*}[t]
\centering
\small
\begin{tabular}{lcccccc}
\hline
Method & Tool Avg. & Math Avg. & Reasoning Avg. & Coding Avg. & Overall Avg. & $\Delta$ \\
\hline
Task Arithmetic & 32.50 & 39.13 & 33.67 & 49.33 & 38.66 & 0.00 \\
Head-only mean & \textbf{42.67} & 29.56 & \underline{33.84} & 46.27 & 38.08 & -0.58 \\
Residual-only mean & \underline{38.65} & 33.73 & 26.94 & \underline{50.94} & 37.56 & -1.10 \\
\rowcolor{oursrowgray}
SRC-A w/o head & 36.81 & \underline{40.38} & 32.83 & 50.81 & \underline{40.21} & +1.55 \\
\rowcolor{oursrowgray}
SRC-A w/ ungated head & 35.36 & 39.76 & 33.33 & 49.57 & 39.50 & +0.84 \\
\rowcolor{oursrowgray}
\textbf{\textsc{ResMerge} (Ours)} & 37.61 & \textbf{41.26} & \textbf{34.34} & \textbf{51.10} & \textbf{41.08} & \textbf{+2.42} \\
\hline
\end{tabular}
\caption{Structural ablation on the Qwen2.5-7B-Base expert group. Category averages are computed by averaging the benchmark-level scores within each capability domain. \textbf{Bold} and \underline{underlined} values denote the best and second-best results in each column. $\Delta$ is measured relative to Task Arithmetic.}
\label{tab:ablation}
\end{table*}

\subsection{Main Results}
\label{sec:main_results}

We present the benchmark-level main comparison on the Qwen2.5-7B-Base expert group in Table~\ref{tab:main_results}, with other expert groups reported in Appendix~\ref{sec:app_full_results}. \textsc{ResMerge} achieves the highest overall average among merged models, reaching 41.08\% and improving over the strongest non-ours baseline, ISO-C, by 2.22 points. It also improves over SRC-A by 0.87 points, indicating that LHC provides additional gains beyond the residual backbone.

Across individual benchmarks, \textsc{ResMerge} achieves the best results on Live P-Mul and MATH500, and remains competitive on tool-use, math, and coding tasks. In contrast, several baselines perform well on specific benchmarks but degrade sharply on others, such as TIES and DARE+TIES on tool-use tasks. This balanced performance leads \textsc{ResMerge} to the best overall result.

\subsection{Ablation Study}
\label{sec:ablation}
We conduct ablation studies on the Qwen2.5-7B-Base expert group to validate the main design choices of our method. These studies help isolate the contribution of each component to the final merged model. The structural ablation examines whether the spectral residual should serve as the backbone and whether the leading head should be injected with reliability control. The hyperparameter ablation studies how the head rank $k$ and injection ratio $\rho$ affect the balance between useful head information and residual backbone stability.
\paragraph{Structural Ablation.}
Table~\ref{tab:ablation} presents structural ablations on the Qwen2.5-7B-Base expert group, with additional ablation results on other expert groups provided in Appendix~\ref{sec:app_full_results}. We include two component-wise averaging baselines: \emph{Head-only mean} averages only the rank-1 head formed by the top singular direction, while \emph{Residual-only mean} averages the residual component after removing this head. We also compare two head-injection variants on top of SRC-A: \emph{SRC-A w/ ungated head} removes the positive agreement gate $S_H$ while keeping the same residual-relative budget, whereas \emph{SRC-A w/ gated head} is our full method.

The results show that simply averaging either component is insufficient. Head-only mean slightly outperforms Residual-only mean in Overall Avg. (38.08\% vs. 37.56\%), suggesting that the head carries useful behavior-specific information. 
However, this does not contradict our residual-backbone design: the head provides sharper task-specific gains, but is less suitable as a stable backbone because such information is concentrated in a narrow spectral direction. 
Although Residual-only mean is weak under naive averaging, SRC-A improves the residual-based backbone from 37.56\% to 40.21\%, indicating that the residual component contains extractable weak consensus requiring a more suitable aggregation rule. 
The key evidence comes from head injection: adding an ungated head to SRC-A decreases Overall Avg. from 40.21\% to 39.50\%, while the gated head improves it to 41.08\%. 
Thus, the head is useful but should not be injected unconditionally. 
With the $S_H$ gate, LHC retains reliable head signals and yields the best overall result.

\paragraph{Design Analysis of Head Rank and Budget.}
We further analyze two key design choices: the head rank $k$ and the residual-relative head budget $\rho$. The former determines how many leading singular directions are treated as the spectral head, while the latter controls the maximum injection strength of LHC relative to the residual backbone. We vary them within a small design-driven range on the Qwen2.5-7B-Base expert group.

\begin{figure}[t]
\centering
\includegraphics[width=\linewidth]{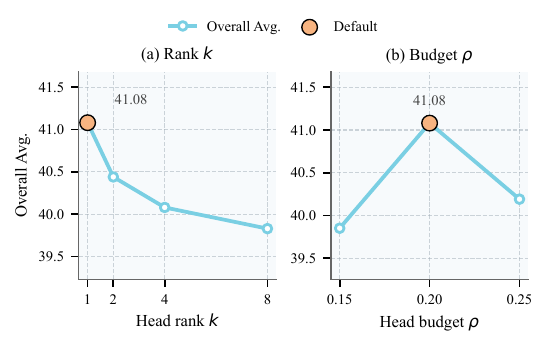}
\caption{Effect of head rank $k$ and residual-relative head budget $\rho$ on the Qwen2.5-7B-Base expert group.}
\label{fig:head_rank_budget}
\end{figure}

For the head rank, increasing $k$ does not improve performance (Figure~\ref{fig:head_rank_budget}(a)). The best overall average is obtained at $k=1$, and performance gradually decreases as more singular directions are included. This suggests that the dominant singular direction benefits most from separate treatment, while unnecessarily enlarging the head may weaken the intended head-residual separation.

For the residual-relative budget, performance follows a clear trade-off (Figure~\ref{fig:head_rank_budget}(b)). A smaller budget underuses useful head information, whereas a larger budget can inject excessive sharp head information into the residual backbone. The middle value $\rho=0.20$ performs best among the tested budgets, supporting a rank-1 head with a moderate and balanced residual-relative budget.

\begin{figure*}[t]
\centering
\includegraphics[width=\textwidth]{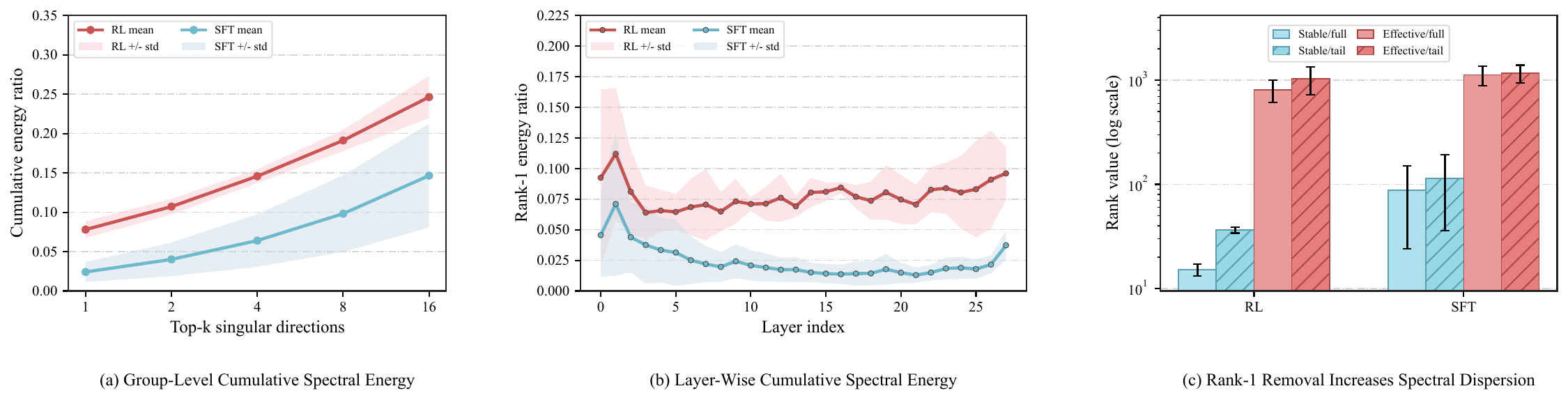}
\caption{
\textbf{Spectral comparison between RL and SFT task vectors. }
RL task vectors show stronger concentration in leading singular directions across groups and layers. 
Removing the leading rank-1 component increases stable and effective ranks, revealing a more dispersed residual component, especially in RL experts.
}
\label{fig:spectral_concentration}
\end{figure*}

\begin{figure}[t]
\centering
\includegraphics[width=1.0\linewidth]{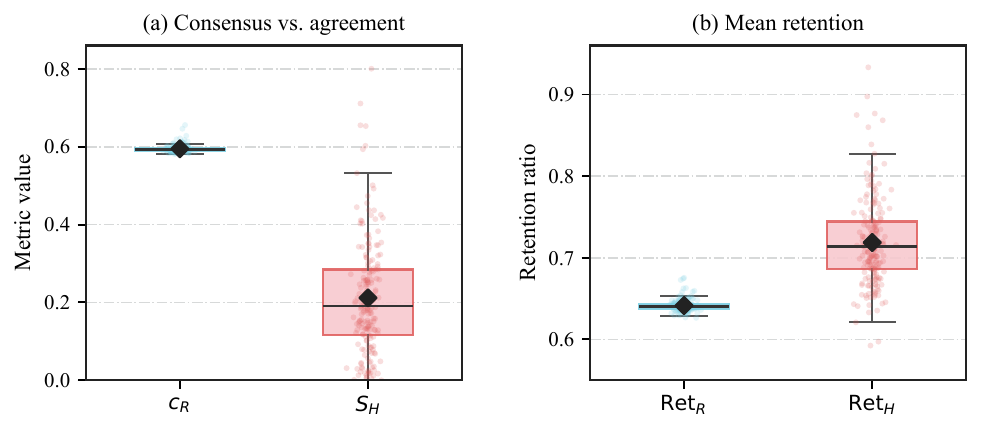}
\caption{\textbf{Geometric consistency analysis of spectral components.} We compare residual spherical coherence, head directional agreement, and mean retention to show that residual components provide a more reliable aggregation backbone while rank-1 heads require reliability-gated lightweight correction.}
\label{fig:geometric_consistency_analysis}
\end{figure}

\subsection{Spectral Structure Analysis}
\label{sec:spectral_structure_analysis}

We compare the singular-value structure of task matrices from RL and SFT post-training. For each mergeable matrix-shaped layer, we compute the SVD and systematically analyze leading-energy concentration, layer-wise spectral concentration, and the resulting change of stable/effective ranks after removing the leading rank-1 component.

Figure~\ref{fig:spectral_concentration}(a) shows that RL task vectors assign more update energy to leading singular directions than SFT task vectors, indicating a less uniform spectrum. Figure~\ref{fig:spectral_concentration}(b) further shows that this spectral concentration consistently appears across layers. These results suggest that a few dominant directions may carry behavior-specific changes and introduce concentrated merging risk if the full task vector is treated as a homogeneous update.

After removing the leading rank-1 component, both stable rank and effective rank increase, especially for RL experts (Figure~\ref{fig:spectral_concentration}(c)). The remaining residual singular directions are distributed across more lower-energy components and are less dominated by any single direction. Together with the ablation results, this supports our component-wise design: SRC-A builds the backbone from dispersed residual singular directions, while LHC injects the sharper leading component only as a lightweight correction controlled by cross-expert agreement.

\subsection{Geometric Consistency Analysis of Spectral Components}
\label{sec:geometric_consistency_analysis}

The spectral analysis above reveals structural differences between leading and residual singular directions, but not their reliability for multi-expert merging. We therefore analyze cross-expert geometric consistency to examine whether the spectral residual forms a stable consensus direction and whether the rank-1 head can be reliably used as a shared lightweight correction.

Figure~\ref{fig:geometric_consistency_analysis}(a) compares two fusion-oriented reliability measures: residual spherical coherence $c_R$ and head agreement $S_H$. These metrics are not directly comparable pairwise similarities: $c_R$ measures whether normalized residual components form a reliable spherical consensus direction, while $S_H$ measures positive directional alignment among rank-1 heads. Residual components show consistently higher spherical coherence, whereas rank-1 heads exhibit weaker and more variable agreement. This supports using the residual component as the SRC-A aggregation backbone and gating the head before adding it through LHC.

We further examine mean retention, which measures how much component norm remains after averaging across experts; its detailed formulation is provided in Appendix~\ref{app:geometry_and_metrics}. Figure~\ref{fig:geometric_consistency_analysis}(b) shows that head retention is higher on average but more variable, while residual retention is lower and more stable. This means that averaging rank-1 heads can still produce a clear average head direction, but this direction is not necessarily reliable when head agreement is weak. In contrast, residual retention provides a conservative scale anchor, while $c_R$ provides evidence for adapting the direction toward a spherical consensus. Together, these observations support the residual-based recomposition strategy: SRC-A builds a stable backbone from the spectral residual, while LHC reintroduces rank-1 head information only as a lightweight correction gated by cross-expert agreement.

\section{Conclusion}

We studied training-free merging of RL post-trained experts and showed that their task vectors exhibit stronger spectral concentration than standard SFT experts. Our analysis reveals that leading and residual singular directions encode complementary behavior knowledge: leading directions are sharp and expressive but sensitive to cross-expert mismatch, while residual directions are more dispersed and provide a stable basis for aggregation. Motivated by this distinction, we proposed \textsc{ResMerge}, which first builds a residual backbone with SRC-A and then injects reliable leading-direction corrections through reliability-gated LHC. Extensive experiments across multiple RL expert groups demonstrate that \textsc{ResMerge} consistently outperforms representative merging baselines, producing stronger, more robust, and more balanced merged models.
\section*{Limitations}

Our study focuses on data-free merging of experts initialized from the same base model.
This controlled setting is useful for isolating task-vector geometry, but it assumes aligned architectures and restricts evaluation to compatible open-source expert groups. 
\textsc{ResMerge} performs static parameter-space merging without input-dependent routing or task-specific calibration, which keeps deployment simple but leaves room for future extensions when task data is available. 
In addition, the current LHC module uses the averaged rank-1 head as the correction direction and controls its magnitude through cross-expert head agreement; more fine-grained or task-aware head composition may further improve merging when expert heads encode highly diverse behaviors. 
Since \textsc{ResMerge} merges RL-trained language model experts in parameter space, the merged model may inherit undesirable behaviors, biases, or safety risks from the source experts. Therefore, merged models should undergo task-specific safety and reliability evaluation before deployment.
Finally, more efficient spectral approximation methods and a formal theoretical understanding of how spectral reliability measures relate to downstream merging performance remain important directions for future work.
\section*{Acknowledgments}

This work was supported by the National Natural Science Foundation of China under Grants U24A20322 and 62576094 and the Beijing Natural Science Foundation (Grant No. L252035).

% Custom bibliography entries only.
\bibliography{main}

\appendix
\clearpage
\section{Appendix}
\label{sec:appendix}

This appendix provides supplementary materials for reproducibility, additional analyses, and extended experimental results. We use publicly available models, benchmarks, and evaluation frameworks following their respective licenses and terms of use.
Appendix~\ref{sec:app_setup} reports model sources, expert-task mappings, datasets and metrics.
Appendix~\ref{sec:app_impl_details} provides additional experimental details.
Appendix~\ref{app:geometry_and_metrics} presents geometric characterizations of ResMerge and defines the diagnostic and evaluation metrics used in our analyses.
Appendix~\ref{sec:app_additional_analysis} presents additional recoverability and component-similarity analyses.
Appendix~\ref{sec:app_full_results} reports full benchmark results and additional ablations.
Appendix~\ref{sec:app_reliability_retention} presents the statistical
reliability and expert-capability retention analyses.

\begin{table*}[t]
\centering
\small
\begin{tabular}{llll}
\toprule
Domain & Dataset & Size & Evaluated Capability \\
\midrule
Coding 
& LiveCodeBench release\_v2
& 511
& Code generation and code reasoning \\

Coding 
& HumanEvalPlus
& 164
& Python programming \\

Coding 
& MBPPPlus
& 378
& Python programming \\

General reasoning
& GPQA-Diamond
& 198
& Graduate-level scientific reasoning \\

Mathematics
& AIME24
& 30
& Competition-style mathematical reasoning \\

Mathematics
& AIME25
& 30
& Competition-style mathematical reasoning \\

Mathematics
& AMC23
& 40
& Competition-style mathematical reasoning \\

Mathematics
& MATH500
& 500
& Mathematical problem solving \\

Tool use
& BFCL V3
& --
& Function and tool calling \\

Memory
& HotpotQA
& 100
& Long-context multi-hop QA \\

Memory
& SQuAD
& 100
& Long-context reading comprehension QA \\
\bottomrule
\end{tabular}
\caption{Dataset details used in our evaluation.}
\label{tab:appendix_datasets}
\end{table*}

\subsection{Model Sources and Evaluation Details}
\label{sec:app_setup}

\textbf{Model sources and expert-task mappings.}
We provide the model sources and expert-task mappings for all expert groups used in our experiments in Table~\ref{tab:appendix_model_sources}. 
Each expert group consists of models that share the same architecture and base initialization, enabling direct weight-space merging. 
For each expert, we report its source link, base model, training paradigm, and corresponding task domain.

\textbf{Datasets and evaluation tools.}
We evaluate across five capability domains: coding, general reasoning, mathematics, tool use, and memory-oriented long-context QA. For coding,
general reasoning, and mathematics benchmarks, we use the open-source
Evalchemy toolkit\footnote{\url{https://github.com/mlfoundations/evalchemy}} for evaluation. For tool-use benchmarks, we use the official BFCL evaluation code\footnote{\url{https://github.com/ShishirPatil/gorilla/tree/main/berkeley-function-call-leaderboard}}, while memory-oriented long-context QA is
evaluated using NeMo-Skills\footnote{\url{https://github.com/NVIDIA-NeMo/Skills}}.
For coding, we use LiveCodeBench release\_v2 to evaluate code generation and code reasoning, and HumanEvalPlus and MBPPPlus to evaluate Python programming. 
For general reasoning, we use GPQA-Diamond, which consists of graduate-level science questions. 
For mathematics, we use AIME24, AIME25, AMC23, and MATH500 to cover competition-style mathematical reasoning and mathematical problem solving. 
For tool use, we use BFCL V3 to evaluate function and tool calling. 
For memory-oriented long-context QA, we sample 100 examples from HotpotQA and 100 examples from SQuAD. 
Table~\ref{tab:appendix_datasets} summarizes dataset sizes and evaluated capabilities.

\textbf{Metrics and aggregation.}
For coding benchmarks, we report Pass@1, which measures the fraction of problems solved by the first generated solution. 
For mathematical reasoning, general reasoning, and memory-oriented QA benchmarks, we report accuracy. 
For tool-use benchmarks, we follow the official BFCL evaluation protocol and report the corresponding BFCL scores for parallel and multiple function-call settings. 
For each capability domain, we first average the benchmark-level scores within that domain to obtain the category average. 
The overall average is then computed as the unweighted mean of all category averages. 
This prevents domains with more benchmarks from dominating the final score.

\begin{table*}[t]
\centering
\scriptsize
\renewcommand{\arraystretch}{1.15}
\setlength{\tabcolsep}{4pt}
\resizebox{\textwidth}{!}{
\begin{tabular}{llll}
\toprule
Model Name / Source & Base Model & Training Paradigm & Expert Domain \\
\midrule
\href{https://huggingface.co/hkust-nlp/Qwen-2.5-7B-SimpleRL-Zoo}{Qwen-2.5-7B-SimpleRL-Zoo}\textbf{ (SimpleRL)} & Qwen2.5-7B-Base & RL post-training & Math \\
\href{https://huggingface.co/Open-Reasoner-Zero/Open-Reasoner-Zero-7B}{Open-Reasoner-Zero-7B}\textbf{ (Zero)} & Qwen2.5-7B-Base & RL post-training & Math reasoning \\
\href{https://huggingface.co/TIGER-Lab/General-Reasoner-Qwen2.5-7B}{General-Reasoner-Qwen2.5-7B}\textbf{ (Reasoner)} & Qwen2.5-7B-Base & RL post-training & General reasoning \\
\href{https://huggingface.co/BillyWang1/qwen2.5-7b-base-retool-slime-sft-v2}
{qwen2.5-7b-base-retool-slime-sft-v2}\textbf{ (ReTool-SFT)}
& Qwen2.5-7B-Base
& Supervised fine-tuning
& Tool use \\

\href{https://huggingface.co/nvidia/AceInstruct-7B}
{AceInstruct-7B}\textbf{ (AceInstruct)}
& Qwen2.5-7B-Base
& Supervised fine-tuning
& Coding \\

\href{https://huggingface.co/OpenDataArena/Qwen2.5-7B-ODA-Math-460k}
{Qwen2.5-7B-ODA-Math-460k}\textbf{ (ODA-Math)}
& Qwen2.5-7B-Base
& Supervised fine-tuning
& Math\\
\midrule
\href{https://huggingface.co/SII-Enigma/Qwen2.5-7B-Ins-AMPO}{Qwen2.5-7B-Ins-AMPO}\textbf{ (AMPO)} & Qwen2.5-7B-Instruct & RL post-training & General reasoning \\
\href{https://huggingface.co/BytedTsinghua-SIA/RL-MemoryAgent-7B}{RL-MemoryAgent-7B}\textbf{ (MemoryAgent)} & Qwen2.5-7B-Instruct & RL post-training & Long-context QA \\
\href{https://huggingface.co/emrecanacikgoz/Qwen2.5-7B-Instruct-ToolRL-grpo-cold}{Qwen2.5-7B-Instruct-ToolRL-grpo-cold}\textbf{ (ToolRL)} & Qwen2.5-7B-Instruct & RL post-training & Tool use \\
\href{https://huggingface.co/Agent-Ark/Toucan-Qwen2.5-7B-Instruct-v0.1}{Toucan-Qwen2.5-7B-Instruct-v0.1} & Qwen2.5-7B-Instruct & Supervised fine-tuning & Tool use \\
\href{https://huggingface.co/stabletoolbench/MirrorAPI}{MirrorAPI} & Qwen2.5-7B-Instruct & Supervised fine-tuning & Tool use \\
\href{https://huggingface.co/open-thoughts/OpenThinker3-7B}{OpenThinker3-7B} & Qwen2.5-7B-Instruct & Supervised fine-tuning & Math \\
\href{https://huggingface.co/bespokelabs/Bespoke-Stratos-7B}{Bespoke-Stratos-7B} & Qwen2.5-7B-Instruct & Supervised fine-tuning & Math reasoning \\
\midrule
\href{https://huggingface.co/lllyx/Qwen3-4B-Base-GRPO}{Qwen3-4B-Base-GRPO}\textbf{ (GRPO)} & Qwen3-4B-Base & RL post-training & Math reasoning \\
\href{https://huggingface.co/TIGER-Lab/General-Reasoner-Qwen3-4B}{General-Reasoner-Qwen3-4B}\textbf{ (General-Reasoner)} & Qwen3-4B-Base & RL post-training & General reasoning \\
\href{https://huggingface.co/spiral-rl/Spiral-Qwen3-4B}{Spiral-Qwen3-4B}\textbf{ (Spiral)} & Qwen3-4B-Base & RL post-training & Reasoning \\
\href{https://huggingface.co/winglian/qwen3-4b-math}{winglian/qwen3-4b-math} & Qwen3-4B-Base & Supervised fine-tuning & Math reasoning \\
\midrule
\href{https://huggingface.co/chengq9/ToolRL-Llama3.2-3B}
{ToolRL-Llama3.2-3B}\textbf{ (ToolRL-llama)}
& Llama-3.2-3B-Instruct
& RL post-training
& Tool use \\

\href{https://huggingface.co/sunblaze-ucb/Llama-3.2-3B-Instruct-GRPO-MATH-1EPOCH}
{Llama-3.2-3B-Instruct-GRPO-MATH-1EPOCH}\textbf{ (GRPO-MATH)}
& Llama-3.2-3B-Instruct
& RL post-training
& Math\\

\href{https://huggingface.co/Menlo/ReZero-v0.1-llama-3.2-3b-it-grpo-250404}
{ReZero-v0.1-llama-3.2-3b-it-grpo-250404}\textbf{ (ReZero)}
& Llama-3.2-3B-Instruct
& RL post-training
& General reasoning \\

\midrule
\href{https://huggingface.co/SII-Enigma/Qwen2.5-7B-Ins-SFT-GRPO}
{Qwen2.5-7B-Ins-SFT-GRPO}\textbf{ (SFT-GRPO)}
& Qwen2.5-7B-Instruct
& SFT followed by RL
& Math\\

\href{https://huggingface.co/andrewlngdn/dsl-debug-7b-sft-rl}
{dsl-debug-7b-sft-rl}\textbf{ (DSL-Debug)}
& Qwen2.5-7B-Instruct
& SFT followed by RL
& Coding \\

\href{https://huggingface.co/xxwu/Agent-STAR-RL-7B}
{Agent-STAR-RL-7B}\textbf{ (Agent-STAR)}
& Qwen2.5-7B-Instruct
& SFT followed by RL
& Tool use \\
\bottomrule
\end{tabular}
}
\caption{Detailed information on the expert models used in our experiments, including their source models, base models, training paradigms, and expert domains.}
\label{tab:appendix_model_sources}
\end{table*}

\subsection{Additional Experimental Details}
\label{sec:app_impl_details}

All merging and evaluation procedures use fixed, deterministic configurations throughout our experiments.

\textbf{Aligned tuning protocol for baselines.}
To ensure a fair comparison, we tune the hyperparameters governing the
merging procedure for the principal baselines using a shared protocol.
Table~\ref{tab:baseline_hparam_search} lists the tuned parameters,
search grids, and selected values. For each baseline, we first identify
the best-performing configuration on the Qwen2.5-7B-Base expert group
according to the overall average score. We then transfer this
configuration unchanged to the remaining expert groups and compare it
against the method's original default configuration. To avoid
underestimating baseline performance, we report, for each expert group,
the configuration that achieves the higher overall average score. This
protocol involves no target-group-specific tuning, task-specific
training data, or gradient-based optimization.

\begin{table*}[t]
\centering
\small
\setlength{\tabcolsep}{10pt}
\renewcommand{\arraystretch}{1.15}

\begin{tabular}{@{}lll c@{}}
\toprule
\textbf{Method}
& \textbf{Tuned Parameters}
& \textbf{Search Grid}
& \textbf{Selected} \\
\midrule
TA
& task-vector scale
& $\{0.7, 1.0, 1.3\}$
& $1.0$ \\

TIES
& density; merge scale
& $\{0.2, 0.5, 0.8\}$; $\{0.7, 1.0, 1.3\}$
& $0.2$; $1.0$ \\

DARE + TIES
& drop rate; merge scale
& $\{0.2, 0.5, 0.8\}$; $\{0.7, 1.0, 1.3\}$
& $0.5$; $0.7$ \\

RAM
& rescale factor
& $\{1.00, 1.05, 1.10, 1.20\}$
& $1.00$ \\

TSV-Merge
& merge scale
& $\{0.7, 1.0, 1.3\}$
& $0.7$ \\
\bottomrule
\end{tabular}

\caption{Hyperparameter search protocol for the baselines
(selected on Qwen2.5-7B-Base).}
\label{tab:baseline_hparam_search}
\end{table*}

\textbf{Sensitivity analysis of $\beta_{\min}$ and $\beta_{\max}$.}
We evaluate all combinations of
$\beta_{\min}\in\{0.6,0.7,0.8\}$ and
$\beta_{\max}\in\{0.8,0.9,1.0\}$ on the
Qwen2.5-7B-Base expert group, while keeping all other settings fixed.
As shown in Table~\ref{tab:beta_sensitivity}, the overall average ranges
from 40.40 to 41.38 across the nine configurations, with a total variation
of less than one point. These results indicate that ResMerge remains
relatively stable under moderate changes to $\beta_{\min}$ and
$\beta_{\max}$ within the evaluated range. Notably, the main results use
the predefined default setting $(0.7,1.0)$ rather than the best-performing
configuration identified in this sensitivity analysis.

\begin{table}[t]
\centering
\small
\renewcommand{\arraystretch}{1.08}
\setlength{\tabcolsep}{9pt}
\begin{tabular}{@{}cccc@{}}
\toprule
\textbf{$\beta_{\min}$}
& \textbf{$\beta_{\max}$}
& \textbf{Overall Avg.}
& \textbf{$\Delta$} \\
\midrule
0.6 & 0.8 & 40.48 & $-0.60$ \\
0.6 & 0.9 & 40.60 & $-0.48$ \\
0.6 & 1.0 & 40.84 & $-0.24$ \\
0.7 & 0.8 & 40.43 & $-0.65$ \\
0.7 & 0.9 & 40.62 & $-0.46$ \\
\textbf{0.7} & \textbf{1.0} & \textbf{41.08} & \textbf{0.00} \\
0.8 & 0.8 & 41.38 & $+0.30$ \\
0.8 & 0.9 & 40.40 & $-0.68$ \\
0.8 & 1.0 & 40.81 & $-0.27$ \\
\bottomrule
\end{tabular}
\caption{Sensitivity of $\beta_{\min}$ and $\beta_{\max}$ on the
Qwen2.5-7B-Base expert group. $\Delta$ is measured relative to the
default setting, shown in bold.}
\label{tab:beta_sensitivity}
\end{table}

\textbf{Computational and memory overhead.}
We evaluate the offline merging efficiency of ResMerge on the
Qwen2.5-7B-Base expert group with three experts. All methods are measured
under the same hardware, implementation pipeline, and streaming setting.
Table~\ref{tab:merging_overhead} reports the end-to-end merge time, the
fraction of merge time spent on SVD operations, the number of SVD calls,
and peak GPU memory.

ResMerge completes merging in $2.22$ minutes, compared with $3.65$ minutes
for TSV-Merge and $3.23$ minutes for ISO-CTS. It also performs fewer SVD
calls, requiring $588$ calls versus $990$ for TSV-Merge and $1188$ for
ISO-CTS. Under the same streaming implementation, ResMerge reaches a peak
GPU memory usage of $8.13$ GB, matching the SVD-free TA baseline and
remaining substantially below the two spectral baselines. The additional
cost relative to TA primarily arises from spectral decomposition.
Nevertheless, model merging is performed only once offline and introduces
no additional inference-time computation or memory overhead. These results
indicate that the layer-wise SVD overhead of ResMerge remains moderate
under the evaluated setting.

\begin{table*}[t]
\centering
\small
\renewcommand{\arraystretch}{1.12}
\setlength{\tabcolsep}{9pt}
\begin{tabular}{lcccc}
\toprule
\textbf{Method}
& \textbf{Merge Time (min)}
& \textbf{SVD Share (\%)}
& \textbf{SVD Calls}
& \textbf{Peak GPU Memory (GB)} \\
\midrule
TA
& $0.34 \pm 0.02$
& N/A
& 0
& 8.13 \\

TSV-Merge
& $3.65 \pm 0.00$
& $96.10 \pm 0.13$
& 990
& 14.46 \\

ISO-CTS
& $3.23 \pm 0.02$
& $95.12 \pm 0.27$
& 1188
& 24.69 \\

\textbf{ResMerge (Ours)}
& $\mathbf{2.22 \pm 0.01}$
& $\mathbf{93.03 \pm 0.15}$
& \textbf{588}
& \textbf{8.13} \\
\bottomrule
\end{tabular}

\caption{
Computational and memory overhead on the Qwen2.5-7B-Base expert group
with three experts, measured using 8$\times$ NVIDIA A6000 GPUs.
SVD Share denotes the proportion of the measured merge time spent on
SVD operations, aggregated across GPUs.
}
\label{tab:merging_overhead}
\end{table*}

\subsection{Geometric Characterization and Metric Definitions}
\label{app:geometry_and_metrics}

This section first provides a geometric characterization of the
head--residual decomposition and the two components of ResMerge, and then
defines the diagnostic and evaluation metrics used in our analyses.

\paragraph{Geometric characterization.}
For clarity, we omit the layer index and the numerical stabilizer
$\epsilon$ in the following derivations.

\emph{Orthogonal head--residual decomposition.}
For expert $i$, the task vector is decomposed into a leading spectral
head $H_i$ and a residual component $R_i$:
\begin{equation}
\Delta_i = H_i + R_i,
\qquad
\langle H_i, R_i \rangle_F = 0.
\label{eq:orthogonal_decomposition}
\end{equation}
The orthogonality follows directly from the truncated singular value
decomposition, since $H_i$ contains the selected leading singular
components and $R_i$ contains the remaining components. Consequently,
\begin{equation}
\|\Delta_i\|_F^2
=
\|H_i\|_F^2+\|R_i\|_F^2.
\label{eq:orthogonal_energy}
\end{equation}
The decomposition therefore partitions the task-vector energy into
non-overlapping spectral components, rather than duplicating or
heuristically discarding update directions.

\emph{Spherical consensus interpretation of SRC-A.}
Let
\begin{equation}
q_i
=
\frac{R_i}{\|R_i\|_F}
\label{eq:normalized_residual}
\end{equation}
denote the normalized residual direction of expert $i$. SRC-A constructs
the aggregate direction
\begin{equation}
q_{\mathrm{src}}
=
\frac{\sum_{i=1}^{N}q_i}
{\left\|\sum_{i=1}^{N}q_i\right\|_F}.
\label{eq:src_consensus}
\end{equation}
Provided that $\sum_i q_i\neq 0$, this direction solves
\begin{equation}
q_{\mathrm{src}}
\in
\operatorname*{arg\,max}_{\|q\|_F=1}
\frac{1}{N}
\sum_{i=1}^{N}
\langle q,q_i\rangle_F.
\label{eq:src_optimization}
\end{equation}
To see this, the objective can be rewritten as
\begin{equation}
\frac{1}{N}
\left\langle
q,\sum_{i=1}^{N}q_i
\right\rangle_F.
\label{eq:src_objective_rewrite}
\end{equation}
Under the constraint $\|q\|_F=1$, the Cauchy--Schwarz inequality gives
\begin{equation}
\left\langle q,\sum_i q_i\right\rangle_F
\leq
\left\|\sum_i q_i\right\|_F.
\label{eq:src_cauchy}
\end{equation}
The upper bound is attained when $q$ is aligned with $\sum_i q_i$,
yielding Eq.~\ref{eq:src_consensus}. Therefore, SRC-A selects the
unit-Frobenius direction that maximizes the average alignment with the
normalized expert residuals. It can thus be interpreted as a spherical
consensus estimator rather than a direct Euclidean average of residual
magnitudes.

\emph{Bounded perturbation interpretation of LHC.}
Let $A$ denote the residual backbone and let
\begin{equation}
\widehat{H}_{\mathrm{mean}}
=
\frac{H_{\mathrm{mean}}}
{\|H_{\mathrm{mean}}\|_F+\epsilon}
\label{eq:normalized_mean_head}
\end{equation}
denote the normalized candidate head direction. The LHC correction is
defined as
\begin{equation}
C
=
\rho S_H\|A\|_F\widehat{H}_{\mathrm{mean}},
\label{eq:lhc_correction}
\end{equation}
where $S_H\in[0,1]$ denotes positive cross-expert head agreement and
$\rho$ controls the maximum injection ratio. Since
$\|\widehat{H}_{\mathrm{mean}}\|_F\leq 1$
and $S_H\in[0,1]$, Eq.~\ref{eq:lhc_correction} implies
\begin{equation}
\|C\|_F
\leq
\rho S_H\|A\|_F
\leq
\rho\|A\|_F.
\label{eq:lhc_bound}
\end{equation}
Applying the triangle inequality together with
Eq.~\ref{eq:lhc_bound} gives
\begin{equation}
\|A+C\|_F
\leq
(1+\rho)\|A\|_F.
\label{eq:merged_update_bound}
\end{equation}
LHC therefore introduces the leading spectral head as an
agreement-aware correction whose magnitude is explicitly bounded
relative to the residual backbone. In particular, when $\rho\leq 1$,
the correction norm cannot exceed that of the backbone.

\paragraph{Metric definitions.}

\emph{Component norm retention.}
For a component $X_i\in\{H_i,R_i\}$, we measure how much component norm
is retained after Euclidean averaging using
\begin{equation}
\operatorname{Ret}_{X}
=
\frac{
\left\|
\frac{1}{N}\sum_{i=1}^{N}X_i
\right\|_F
}{
\frac{1}{N}\sum_{i=1}^{N}\|X_i\|_F
}.
\label{eq:component_retention}
\end{equation}
By the triangle inequality,
$\operatorname{Ret}_{X}\in[0,1]$. A higher value indicates less norm loss
due to directional cancellation. We report this metric separately for
the spectral heads and residuals.

\emph{Component performance recovery.}
For each expert $E_i$, we evaluate the full task vector and its
component-only variants on the benchmark suite used for the expert group
sharing the same base model as $E_i$. Let $\mathcal{D}_i$ denote the set
of capability domains in this benchmark suite, let
$\mathcal{B}_{i,d}$ denote the benchmarks in domain $d$, and let
$s_b(M)$ denote the score of model $M$ on benchmark $b$. Following the
same aggregation rule as the Overall Avg. in the main experiments, we
define
\begin{equation}
S_i(M)
=
\frac{1}{|\mathcal{D}_i|}
\sum_{d\in\mathcal{D}_i}
\frac{1}{|\mathcal{B}_{i,d}|}
\sum_{b\in\mathcal{B}_{i,d}}
s_b(M).
\label{eq:component_recovery_score}
\end{equation}
This aggregation first averages benchmark scores within each capability
domain and then averages the resulting domain scores equally.

\begin{figure}[t]
\centering
\includegraphics[width=\linewidth]{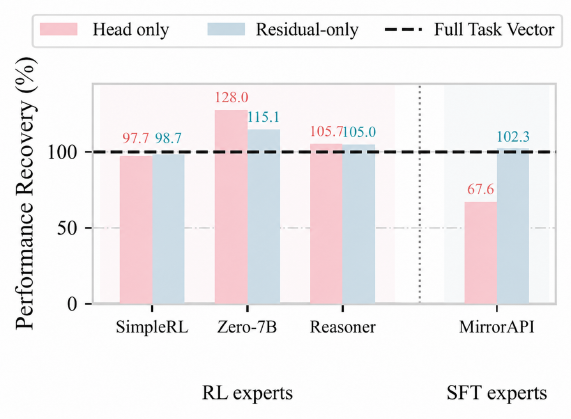}
\caption{
Additional component-level recovery results for RL and SFT post-training experts.
We compare full task vectors with head-only and residual-only variants. 
The additional results further show that RL task vectors often preserve non-trivial behavior knowledge in both spectral components, while SFT experts exhibit more uneven component recoverability.
}
\label{fig:app_recovery_more}
\end{figure}

For a component $X_i\in\{H_i,R_i\}$, its performance recovery relative
to the full task vector is defined as
\begin{equation}
\operatorname{Rec}_{X_i}
=
100 \times
\frac{
S_i(W_0+X_i)
}{
S_i(W_0+\Delta_i)
}.
\label{eq:component_recovery}
\end{equation}
Since $W_0+\Delta_i=W_i$, the full task vector has a recovery value of
$100\%$. Figure~\ref{fig:app_recovery_more} reports this normalized
recovery directly, while Figure~\ref{fig:motivation} uses the
corresponding centered form
\begin{equation}
\operatorname{CRec}_{X_i}
=
\operatorname{Rec}_{X_i}-100,
\label{eq:centered_component_recovery}
\end{equation}
for which the full task vector is centered at $0\%$. Positive values
indicate that the component-only model exceeds the full task-vector
performance, whereas negative values indicate partial recovery.

\emph{Capability-retention metrics.}
Let $\mathcal{C}$ denote the set of evaluated capability domains, and
let $s(M,c)$ denote the category-average score of model $M$ on
capability $c\in\mathcal{C}$. For each capability, we use the strongest
source expert as the reference:
\begin{equation}
s^{\star}(c)
=
\max_{1\leq i\leq N}s(E_i,c).
\end{equation}
The expert-capability retention of a merged model $M$ on capability $c$
is defined as
\begin{equation}
\operatorname{ECR}_{c}(M)
=
\frac{s(M,c)}{s^{\star}(c)}.
\label{eq:ecr}
\end{equation}
An ECR of $1$ indicates that the merged model matches the strongest
source expert on that capability, while a value above $1$ indicates
that it exceeds the strongest source-expert performance.

We report \emph{Mean ECR} and \emph{Min ECR}, defined respectively as
the average and minimum ECR across capability domains, to characterize
average and worst-case relative capability retention.

To complement these relative measures, the worst capability shortfall is
defined as
\begin{equation}
\operatorname{WCS}(M)
=
\max_{c\in\mathcal{C}}
\left[
s^{\star}(c)-s(M,c)
\right]_{+},
\label{eq:wcs}
\end{equation}
where $[x]_{+}=\max(x,0)$. WCS measures the largest absolute performance
drop relative to the strongest source expert on the same capability.
Lower values indicate stronger worst-case capability preservation.

\emph{Benchmark-composition robustness.}
We further evaluate whether the improvement of \textsc{ResMerge} depends
disproportionately on a small number of benchmarks. For benchmark $b$,
we define the score difference relative to a fixed comparison baseline
$M^{\star}$ as
\begin{equation}
\delta_b
=
s_b(\textsc{ResMerge})
-
s_b(M^{\star}),
\label{eq:benchmark_delta}
\end{equation}
where $M^{\star}$ is the strongest non-ours baseline for the
corresponding expert group.

Following the same domain-balanced aggregation rule above, the overall
difference $\Delta$ is obtained by first averaging benchmark-level
differences within each capability domain and then averaging equally
across domains. Thus, each capability domain contributes equally,
regardless of the number of benchmarks it contains.

For the domain-stratified benchmark bootstrap, we sample benchmarks with
replacement independently within each domain while preserving the
original number of benchmarks in that domain. We recompute $\Delta$ for
each bootstrap replicate and report the $95\%$ percentile bootstrap
interval of the resulting distribution, together with the proportion of
replicates for which $\Delta>0$. The latter is an empirical bootstrap
positivity rate rather than a conventional hypothesis-test $p$-value.

For the leave-one-benchmark-out analysis, we remove each benchmark once
and recompute the same domain-aware $\Delta$. If the removed benchmark is
the only entry in its domain, the resulting empty domain is excluded from
that leave-one-out average. We report the number of leave-one-out
settings with a positive $\Delta$ and the minimum observed
leave-one-out $\Delta$. Together, these summaries assess the robustness
of the observed improvement to changes in benchmark composition.

\begin{figure*}[t]
\centering
\includegraphics[width=0.95\textwidth]{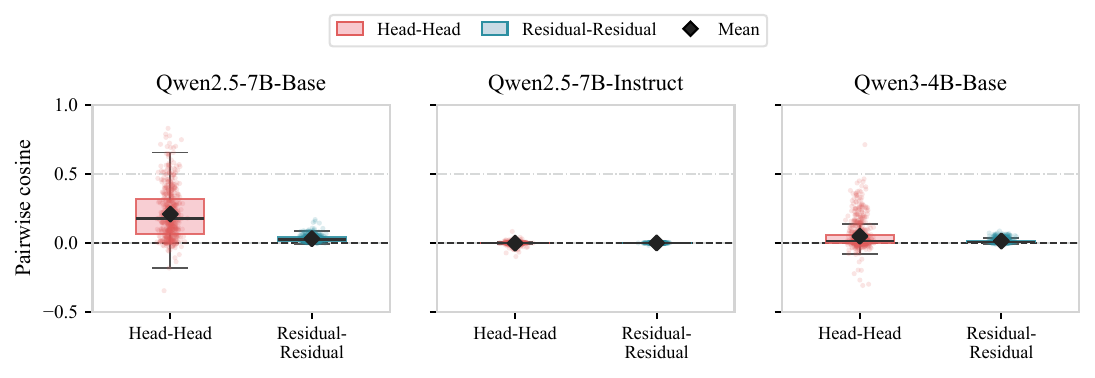}
\caption{
Pairwise similarity of rank-1 heads and residual components across experts.
Rank-1 heads show higher average pairwise cosine but also larger variance, suggesting useful yet unstable cross-expert alignment. 
Residual components exhibit low pairwise cosine, consistent with their high-dimensional dispersed spectrum. 
Thus, residual reliability should be understood through aggregate spherical coherence rather than pairwise similarity.
}
\label{fig:app_head_residual_similarity}
\end{figure*}

\begin{table*}[t]
\centering
\scriptsize
\renewcommand{\arraystretch}{1.25}
\setlength{\tabcolsep}{3pt}
\resizebox{\textwidth}{!}{
\begin{tabular}{lccccccccccccc}
\toprule
\multicolumn{1}{c}{Method}
& \multicolumn{4}{c}{Tool Using}
& \multicolumn{4}{c}{Math}
& Reasoning
& \multicolumn{3}{c}{Coding}
& Overall \\
\cmidrule(lr){2-5}
\cmidrule(lr){6-9}
\cmidrule(lr){10-10}
\cmidrule(lr){11-13}
\cmidrule(lr){14-14}
&
Live Para & Live P-Mul & Non-Live Para & Non-Live P-Mul
& AIME24 & AIME25 & AMC23 & MATH500
& GPQA-D
& LiveCodeBench & HumanEval++ & MBPP++
& Overall Avg. \\
\midrule
\rowcolor{groupgray}
\multicolumn{14}{c}{\textit{Base and Expert Models}} \\
\textbf{Qwen3-4B} & 50.00 & 25.00 & 56.50 & 28.00 & 13.33 & 6.67 & 32.50 & 52.40 & 19.70 & 4.70 & 77.44 & 5.03 & 28.71 \\
\textbf{GRPO} & 31.25 & 29.17 & 68.50 & 72.00 & 23.33 & 16.67 & 72.50 & 77.40 & 31.48 & 5.28 & 75.00 & 55.03 & 43.57 \\
\textbf{General-Reasoner} & 68.75 & 58.33 & 78.50 & 81.50 & 26.67 & 16.67 & 45.00 & 66.00 & 36.87 & 18.40 & 81.71 & 55.29 & 49.76 \\
\textbf{Spiral} & 6.25 & 12.50 & 57.00 & 71.50 & 13.33 & 16.67 & 42.50 & 73.00 & 29.29 & 9.98 & 76.22 & 65.87 & 38.29 \\
\midrule
\rowcolor{groupgray}
\multicolumn{14}{c}{\textit{Merged Models}} \\
\textbf{TA} & \firstscore{56.25} & \firstscore{37.50} & \firstscore{76.50} & \firstscore{75.50} & \thirdscore{20.00} & \thirdscore{13.33} & \thirdscore{55.00} & 76.00 & 31.65 & 9.59 & 76.22 & 57.14 & \thirdscore{45.46} \\
\textbf{TIES} & 6.25 & 16.67 & 43.50 & 55.50 & \secondscore{23.33} & \firstscore{20.00} & \secondscore{65.00} & \thirdscore{77.80} & \secondscore{34.34} & \firstscore{15.66} & \firstscore{81.71} & \secondscore{59.79} & 40.93 \\
\textbf{DARE + TIES} & \thirdscore{12.50} & \thirdscore{20.83} & 44.50 & 61.50 & \firstscore{26.67} & \secondscore{16.67} & \secondscore{65.00} & \thirdscore{77.80} & \thirdscore{34.18} & 9.98 & \secondscore{80.49} & 59.26 & 41.36 \\
\textbf{TSV-Merge} & \secondscore{50.00} & \secondscore{33.33} & \thirdscore{67.50} & \thirdscore{71.50} & \thirdscore{20.00} & \secondscore{16.67} & \secondscore{65.00} & 76.00 & 32.15 & 10.57 & \thirdscore{79.27} & \thirdscore{59.52} & \secondscore{45.48} \\
\textbf{RAM} & \thirdscore{12.50} & \thirdscore{20.83} & 44.50 & 61.50 & \secondscore{23.33} & \secondscore{16.67} & \firstscore{70.00} & \firstscore{78.40} & \firstscore{34.51} & \secondscore{14.48} & \firstscore{81.71} & \firstscore{61.11} & 42.22 \\
\midrule
\textbf{\textsc{ResMerge} (Ours)} & \secondscore{50.00} & \firstscore{37.50} & \secondscore{74.50} & \secondscore{72.50} & \secondscore{23.33} & \firstscore{20.00} & \secondscore{65.00} & \secondscore{78.00} & 33.67 & \thirdscore{13.50} & 78.66 & 58.20 & \firstscore{47.25} \\
\bottomrule
\end{tabular}
}
\caption{Main results on the Qwen3-4B-Base expert group across individual benchmarks and the overall average. Shaded cells denote the top distinct results among merged models in each column: \colorbox{rankfirst}{\textbf{1st}}, \colorbox{ranksecond}{2nd}, and \colorbox{rankthird}{3rd}; tied scores share the same color.}
\label{tab:main_results_qwen3_4b}
\end{table*}

\begin{table*}[t]
\centering
\small
\begin{tabular}{lcccccc}
\hline
Method & Tool Avg. & Math Avg. & Reasoning Avg. & Coding Avg. & Overall Avg. & $\Delta$ \\
\hline
Task Arithmetic & \underline{61.44} & 41.08 & 31.65 & 47.65 & 45.46 & 0.00 \\
Head-only mean & 42.71 & 33.58 & 23.40 & 32.61 & 33.07 & -12.39 \\
Residual-only mean & 58.38 & 43.58 & 31.99 & \underline{49.50} & 45.86 & +0.40 \\
\rowcolor{oursrowgray}
SRC-A w/o head & 58.75 & \underline{45.44} & \underline{32.66} & 49.49 & \underline{46.59} & \underline{+1.13} \\
\rowcolor{oursrowgray}
SRC-A w/ ungated head & \textbf{63.52} & 41.49 & 32.15 & 47.06 & 46.05 & +0.59 \\
\rowcolor{oursrowgray}
\textbf{\textsc{ResMerge} (Ours)} & 58.63 & \textbf{46.58} & \textbf{33.67} & \textbf{50.12} & \textbf{47.25} & \textbf{+1.79} \\
\hline
\end{tabular}
\caption{Structural ablation on the Qwen3-4B-Base expert group. Category averages are computed by averaging the benchmark-level scores within each capability domain. \textbf{Bold} and \underline{underlined} values denote the best and second-best results in each column. $\Delta$ is measured relative to Task Arithmetic.}
\label{tab:ablation_qwen3_4b}
\end{table*}

\begin{table*}[t]
\centering
\scriptsize
\renewcommand{\arraystretch}{1.25}
\setlength{\tabcolsep}{3pt}
\resizebox{\textwidth}{!}{
\begin{tabular}{lccccccccccccc}
\toprule
\multicolumn{1}{c}{Method}
& \multicolumn{4}{c}{Tool Using}
& \multicolumn{4}{c}{Math}
& Reasoning
& \multicolumn{3}{c}{Memory}
& Overall \\
\cmidrule(lr){2-5}
\cmidrule(lr){6-9}
\cmidrule(lr){10-10}
\cmidrule(lr){11-13}
\cmidrule(lr){14-14}
&
Live Para & Live P-Mul & Non-Live Para & Non-Live P-Mul
& AIME24 & AIME25 & AMC23 & MATH500
& GPQA-D
& HotpotQA-7K & HotpotQA-14K & SQuAD-32K
& Overall Avg. \\
\midrule
\rowcolor{groupgray}
\multicolumn{14}{c}{\textit{Base and Expert Models}} \\
\textbf{Qwen2.5-7B-Instruct} & 62.50 & 66.67 & 91.50 & 82.50 & 13.33 & 0.00 & 50.00 & 72.40 & 23.23 & 60.00 & 59.00 & 72.00 & 49.16 \\
\textbf{AMPO} & 68.75 & 50.00 & 92.00 & 84.00 & 13.33 & 13.33 & 55.00 & 76.20 & 30.13 & 64.00 & 61.00 & 75.00 & 52.49 \\
\textbf{MemoryAgent} & 68.75 & 66.67 & 89.50 & 85.00 & 10.00 & 3.33 & 45.00 & 71.20 & 17.67 & 65.00 & 64.00 & 78.00 & 49.13 \\
\textbf{ToolRL} & 68.75 & 70.83 & 90.50 & 85.00 & 10.00 & 6.67 & 47.50 & 72.00 & 23.91 & 61.00 & 61.00 & 75.00 & 50.60 \\
\midrule
\rowcolor{groupgray}
\multicolumn{14}{c}{\textit{Merged Models}} \\

\textbf{TA}
& \secondscore{68.75}
& \secondscore{66.67}
& \firstscore{91.00}
& \thirdscore{86.50}
& \secondscore{13.33}
& 3.33
& \secondscore{55.00}
& 74.60
& \secondscore{26.26}
& \thirdscore{64.00}
& 62.00
& 76.00
& 52.10 \\

\textbf{TIES}
& \secondscore{68.75}
& \secondscore{66.67}
& \thirdscore{90.00}
& 86.00
& \firstscore{16.67}
& \thirdscore{10.00}
& \thirdscore{52.50}
& \firstscore{75.60}
& \thirdscore{25.08}
& \secondscore{65.00}
& \thirdscore{63.00}
& \thirdscore{77.00}
& \secondscore{52.49} \\

\textbf{DARE + TIES}
& \thirdscore{62.50}
& \secondscore{66.67}
& 89.00
& \secondscore{87.00}
& \firstscore{16.67}
& \firstscore{20.00}
& \thirdscore{52.50}
& \secondscore{75.20}
& 23.40
& \firstscore{67.00}
& 60.00
& \secondscore{78.00}
& 52.28 \\

\textbf{TSV-Merge}
& \secondscore{68.75}
& \secondscore{66.67}
& \secondscore{90.50}
& \thirdscore{86.50}
& \secondscore{13.33}
& \thirdscore{10.00}
& \secondscore{55.00}
& 74.40
& 24.58
& \thirdscore{64.00}
& \firstscore{65.00}
& \thirdscore{77.00}
& \thirdscore{52.38} \\

\textbf{ISO-C}
& \secondscore{68.75}
& \secondscore{66.67}
& \thirdscore{90.00}
& \firstscore{87.50}
& \secondscore{13.33}
& \secondscore{13.33}
& \thirdscore{52.50}
& 74.60
& 23.57
& \secondscore{65.00}
& 62.00
& \thirdscore{77.00}
& 52.06 \\

\textbf{ISO-CTS}
& \secondscore{68.75}
& \secondscore{66.67}
& \firstscore{91.00}
& \firstscore{87.50}
& \firstscore{16.67}
& \thirdscore{10.00}
& 47.50
& 74.20
& 24.75
& \thirdscore{64.00}
& 61.00
& \secondscore{78.00}
& 52.00 \\

\textbf{RAM}
& \secondscore{68.75}
& \secondscore{66.67}
& \thirdscore{90.00}
& 85.50
& \thirdscore{10.00}
& 6.67
& \firstscore{57.50}
& \thirdscore{74.80}
& 24.58
& 63.00
& \secondscore{64.00}
& \firstscore{80.00}
& 52.14 \\

\midrule

\textbf{\textsc{ResMerge} (Ours)}
& \firstscore{75.00}
& \firstscore{70.83}
& \firstscore{91.00}
& \thirdscore{86.50}
& \firstscore{16.67}
& \secondscore{13.33}
& \secondscore{55.00}
& 73.00
& \firstscore{27.61}
& \secondscore{65.00}
& 60.00
& 76.00
& \firstscore{53.74} \\
\bottomrule
\end{tabular}
}
\caption{Main results on the Qwen2.5-7B-Instruct expert group across individual benchmarks and the overall average. Shaded cells denote the top distinct results among merged models in each column: \colorbox{rankfirst}{\textbf{1st}}, \colorbox{ranksecond}{2nd}, and \colorbox{rankthird}{3rd}; tied scores share the same color.}
\label{tab:main_results_qwen25_instruct}
\end{table*}

\begin{table*}[t]
\centering
\small
\begin{tabular}{lcccccc}
\hline
Method & Tool Avg. & Math Avg. & Reasoning Avg. & Memory Avg. & Overall Avg. & $\Delta$ \\
\hline
Task Arithmetic & 78.23 & 36.57 & 26.26 & 67.33 & 52.10 & 0.00 \\
Head-only mean & 76.17 & 35.83 & 24.24 & 65.00 & 50.31 & -1.79 \\
Residual-only mean & 78.36 & 37.61 & 25.42 & \textbf{68.00} & 52.35 & +0.25 \\
\rowcolor{oursrowgray}
SRC-A w/o head & 79.27 & \underline{38.29} & \textbf{28.11} & \underline{67.67} & \underline{53.33} & \underline{+1.23} \\
\rowcolor{oursrowgray}
SRC-A w/ ungated head & \underline{79.40} & 36.98 & 26.60 & 64.67 & 51.91 & -0.19 \\
\rowcolor{oursrowgray}
\textbf{\textsc{ResMerge} (Ours)} & \textbf{80.83} & \textbf{39.50} & \underline{27.61} & 67.00 & \textbf{53.74} & \textbf{+1.64} \\
\hline
\end{tabular}
\caption{Structural ablation on the Qwen2.5-7B-Instruct expert group. Category averages are computed by averaging the benchmark-level scores within each capability domain. \textbf{Bold} and \underline{underlined} values denote the best and second-best results in each column. $\Delta$ is measured relative to Task Arithmetic.}
\label{tab:ablation_qwen25_instruct}
\end{table*}

\subsection{Additional Analyses}
\label{sec:app_additional_analysis}

In this section, we provide two supplementary analyses to support the observations in the main paper. 
First, we include component-level recovery experiments on additional RL- and SFT-trained experts. 
Second, we analyze the similarity between leading singular directions and residual singular directions. 
These results further examine whether both spectral components retain recoverable behavior knowledge and whether they capture different parts of the task update.

\paragraph{Additional component recovery results.}
Figure~\ref{fig:app_recovery_more} reports component-level recovery results on additional RL and SFT experts. 
For each expert, we compare the full task vector with two component-only variants: \emph{Head-only}, which keeps only the leading rank-1 spectral head, and \emph{Residual-only}, which keeps the residual component after removing the leading head. 
These additional results examine whether the component-level recoverability observed in the main text holds across different expert groups and training paradigms.

\paragraph{Pairwise similarity of spectral components.}
We further report pairwise cosine similarities among rank-1 heads and among residual components across experts. 
As shown in Figure~\ref{fig:app_head_residual_similarity}, rank-1 heads show higher average pairwise cosine than residual components, but their similarities also have much larger variance across tensors and layers. 
This indicates that head directions are not pure noise: they can exhibit meaningful cross-expert alignment in some modules. 
However, this alignment is unstable, which makes direct head averaging unreliable without an agreement-based gate.

In contrast, residual components have low pairwise cosine, which is expected because they are high-dimensional and spectrally dispersed after removing the leading singular direction. 
Therefore, residual reliability should not be interpreted as strong pairwise alignment. 
Rather, as shown by the residual coherence analysis in the main paper, normalized residual components can still form a stable aggregate spherical consensus direction. 
This distinction motivates our design: SRC-A extracts an aggregate residual consensus instead of relying on high pairwise similarity, while LHC gates the head correction according to head agreement.

\subsection{Full Results and Additional Ablations}
\label{sec:app_full_results}

Beyond the primary Qwen2.5-7B-Base RL setting reported in
Table~\ref{tab:main_results}, we extend the evaluation to two additional
Qwen-based RL expert groups and one Llama-based RL expert group. We also
consider mixed SFT-then-RL and SFT-only expert groups, where the task
vectors contain updates from supervised fine-tuning. These experiments
broaden the evaluation beyond the primary RL-only Qwen setting and help
clarify the scope of \textsc{ResMerge}. We further report domain-level
structural ablations on two additional Qwen-based RL expert groups.

\paragraph{Additional Qwen-based RL expert groups.}
Tables~\ref{tab:main_results_qwen3_4b} and
\ref{tab:main_results_qwen25_instruct} report results on the
Qwen3-4B-Base and Qwen2.5-7B-Instruct RL expert groups, respectively.
\textsc{ResMerge} achieves the highest Overall Avg.\ on both groups.
Together with the primary Qwen2.5-7B-Base results, these findings show
consistent gains across three Qwen-based RL settings.

\begin{table*}[t]
\centering
\scriptsize
\renewcommand{\arraystretch}{1.25}
\setlength{\tabcolsep}{3pt}
\resizebox{\textwidth}{!}{
\begin{tabular}{lcccccccccc}
\toprule
\multicolumn{1}{c}{Method}
& \multicolumn{4}{c}{Tool Using}
& \multicolumn{4}{c}{Math}
& Reasoning
& Overall \\
\cmidrule(lr){2-5}
\cmidrule(lr){6-9}
\cmidrule(lr){10-10}
\cmidrule(lr){11-11}
&
Live Para & Live P-Mul & Non-Live Para & Non-Live P-Mul
& AIME24 & AIME25 & AMC23 & MATH500
& GPQA-D
& Overall Avg. \\
\midrule

\rowcolor{groupgray}
\multicolumn{11}{c}{\textit{Base and Expert Models}} \\

\textbf{Llama-3.2-3B-Instruct}
& 37.50 & 12.50 & 44.50 & 24.50
& 6.67 & 0.00 & 20.00 & 42.80
& 24.58
& 23.90 \\

\textbf{ToolRL-llama}
& 37.50 & 16.67 & 48.00 & 21.50
& 3.33 & 0.00 & 20.00 & 40.00
& 19.87
& 22.21 \\

\textbf{GRPO-MATH}
& 31.25 & 4.17 & 33.50 & 9.00
& 13.33 & 0.00 & 22.50 & 44.00
& 19.36
& 19.60 \\

\textbf{ReZero}
& 25.00 & 16.67 & 17.00 & 6.00
& 6.67 & 0.00 & 7.50 & 34.80
& 31.65
& 20.02 \\

\midrule
\rowcolor{groupgray}
\multicolumn{11}{c}{\textit{Merged Models}} \\

\textbf{TA}
& \thirdscore{25.00}
& \firstscore{12.50}
& \secondscore{37.00}
& 15.50
& \thirdscore{3.33}
& \secondscore{0.00}
& \thirdscore{22.50}
& 41.20
& 20.70
& \thirdscore{19.99} \\

\textbf{TIES}
& \thirdscore{0.00}
& \thirdscore{0.00}
& 27.50
& \thirdscore{16.00}
& \thirdscore{3.33}
& \secondscore{0.00}
& 10.00
& 19.40
& 8.92
& 9.33 \\

\textbf{DARE + TIES}
& \thirdscore{0.00}
& \thirdscore{0.00}
& 0.00
& 0.00
& 0.00
& \secondscore{0.00}
& 0.00
& 0.60
& 1.01
& 0.39 \\

\textbf{TSV-Merge}
& \firstscore{37.50}
& \firstscore{12.50}
& 31.00
& 9.50
& \secondscore{6.67}
& \secondscore{0.00}
& 15.00
& \thirdscore{43.60}
& \thirdscore{21.21}
& \secondscore{20.05} \\

\textbf{ISO-C}
& \secondscore{31.25}
& \firstscore{12.50}
& \firstscore{46.50}
& \firstscore{24.00}
& 0.00
& \secondscore{0.00}
& 0.00
& 16.00
& 17.00
& 16.52 \\

\textbf{RAM}
& \firstscore{37.50}
& \secondscore{8.33}
& 10.50
& 4.50
& \secondscore{6.67}
& \firstscore{3.33}
& \secondscore{27.50}
& \secondscore{44.40}
& \firstscore{23.40}
& 19.69 \\

\midrule
\textbf{\textsc{ResMerge} (Ours)}
& \thirdscore{25.00}
& \firstscore{12.50}
& \thirdscore{34.00}
& \secondscore{18.00}
& \firstscore{10.00}
& \secondscore{0.00}
& \firstscore{30.00}
& \firstscore{47.00}
& \secondscore{22.90}
& \firstscore{22.34} \\

\bottomrule
\end{tabular}
}
\caption{
Main results on the Llama-3.2-3B-Instruct expert group across
individual benchmarks and the overall domain-mean average.
Shaded cells denote the top distinct results among merged models
in each column:
\colorbox{rankfirst}{\textbf{1st}},
\colorbox{ranksecond}{2nd}, and
\colorbox{rankthird}{3rd};
tied scores share the same color.
}
\label{tab:main_results_llama32_3b}
\end{table*}

\begin{table*}[t]
\centering
\scriptsize
\renewcommand{\arraystretch}{1.25}
\setlength{\tabcolsep}{3pt}
\resizebox{\textwidth}{!}{
\begin{tabular}{lccccccccccccc}
\toprule
\multicolumn{1}{c}{Method}
& \multicolumn{4}{c}{Tool Using}
& \multicolumn{4}{c}{Math}
& Reasoning
& \multicolumn{3}{c}{Coding}
& Overall \\
\cmidrule(lr){2-5}
\cmidrule(lr){6-9}
\cmidrule(lr){10-10}
\cmidrule(lr){11-13}
\cmidrule(lr){14-14}
&
Live Para
& Live P-Mul
& Non-Live Para
& Non-Live P-Mul
& AIME24
& AIME25
& AMC23
& MATH500
& GPQA-D
& LiveCodeBench
& HumanEval+
& MBPP+
& Overall Avg. \\
\midrule

\rowcolor{groupgray}
\multicolumn{14}{c}{\textit{Base and Expert Models}} \\

\textbf{Qwen2.5-7B-Instruct}
& 62.50
& 66.67
& 91.50
& 82.50
& 13.33
& 0.00
& 50.00
& 72.40
& 23.23
& 33.46
& 76.83
& 66.93
& 48.01 \\

\textbf{SFT-GRPO}
& 62.50
& 50.00
& 83.50
& 83.50
& 16.67
& 13.33
& 60.00
& 74.00
& 35.02
& 20.35
& 75.61
& 69.31
& 50.25 \\

\textbf{DSL-Debug}
& 56.25
& 58.33
& 91.00
& 86.50
& 13.33
& 10.00
& 55.00
& 72.20
& 22.90
& 32.88
& 75.00
& 66.14
& 47.89 \\

\textbf{Agent-STAR}
& 68.75
& 66.67
& 90.50
& 84.00
& 13.33
& 10.00
& 55.00
& 73.60
& 37.21
& 17.42
& 73.78
& 68.52
& 51.48 \\

\midrule
\rowcolor{groupgray}
\multicolumn{14}{c}{\textit{Merged Models}} \\

\textbf{TA}
& \firstscore{68.75}
& \firstscore{58.33}
& \thirdscore{91.00}
& \thirdscore{84.00}
& \secondscore{20.00}
& \firstscore{13.33}
& \firstscore{57.50}
& \secondscore{75.40}
& \thirdscore{31.82}
& \firstscore{32.29}
& \firstscore{78.05}
& \secondscore{68.25}
& \secondscore{52.11} \\

\textbf{TSV-Merge}
& \secondscore{62.50}
& \firstscore{58.33}
& \secondscore{91.50}
& \firstscore{85.00}
& \secondscore{20.00}
& \secondscore{10.00}
& \firstscore{57.50}
& \firstscore{76.20}
& \firstscore{35.52}
& \thirdscore{29.55}
& \secondscore{77.44}
& \thirdscore{67.20}
& \firstscore{52.21} \\

\midrule
\textbf{\textsc{ResMerge} (Ours)}
& \firstscore{68.75}
& \firstscore{58.33}
& \firstscore{92.00}
& \secondscore{84.50}
& \firstscore{23.33}
& \thirdscore{6.67}
& \secondscore{52.50}
& \thirdscore{74.60}
& \secondscore{32.15}
& \secondscore{30.53}
& \firstscore{78.05}
& \firstscore{68.52}
& \thirdscore{51.59} \\

\bottomrule
\end{tabular}
}
\caption{
Model merging results on the mixed SFT--RL expert group across
individual benchmarks and the overall average. All experts share
Qwen2.5-7B-Instruct as the base model and undergo task-specific SFT
followed by RL. Shaded cells denote the top distinct results among
merged models in each column:
\colorbox{rankfirst}{\textbf{1st}},
\colorbox{ranksecond}{2nd}, and
\colorbox{rankthird}{3rd}; tied scores share the same color.
}
\label{tab:main_results_sft_rl}
\end{table*}

\paragraph{Cross-backbone evaluation on Llama RL experts.}
Table~\ref{tab:main_results_llama32_3b} reports results on three RL experts
derived from Llama-3.2-3B-Instruct. \textsc{ResMerge} achieves the highest
Overall Avg.\ of 22.34, exceeding TSV-Merge and TA by 2.29 and 2.35
points, respectively. It also obtains the highest Math Avg. These results
support the applicability of the residual-based spectral design beyond
the Qwen backbone family in the RL setting.

\paragraph{Mixed SFT--RL expert group.}
We next consider the common SFT-then-RL pipeline, in which each expert
undergoes task-specific supervised fine-tuning followed by reinforcement
learning:
$W_0 \rightarrow W_{\mathrm{SFT}}^{(i)}
\rightarrow W_{\mathrm{SFT+RL}}^{(i)}$.
Accordingly, the task vector
$\Delta_i=W_{\mathrm{SFT+RL}}^{(i)}-W_0$
contains updates introduced by both stages.

As shown in Table~\ref{tab:main_results_sft_rl}, \textsc{ResMerge}
achieves or ties the best score on six of the twelve benchmarks, including
the highest scores on Non-Live Para, AIME24, and MBPP+. Its Overall Avg.\
is 51.59, only 0.62 points below the best result of 52.21 achieved by
TSV-Merge. These results show that \textsc{ResMerge} remains competitive
under the practical SFT-then-RL pipeline, but no longer exhibits the
consistent overall advantage observed in the RL-only settings. This
attenuation is consistent with the RL-specific motivation of the method:
mixing SFT updates into the task vectors may weaken the head--residual
structure exploited by \textsc{ResMerge}.

\begin{table*}[t]
\centering
\scriptsize
\renewcommand{\arraystretch}{1.25}
\setlength{\tabcolsep}{3pt}
\resizebox{\textwidth}{!}{
\begin{tabular}{lccccccccccccc}
\toprule
\multicolumn{1}{c}{Method}
& \multicolumn{4}{c}{Tool Using}
& \multicolumn{4}{c}{Math}
& Reasoning
& \multicolumn{3}{c}{Coding}
& Overall \\
\cmidrule(lr){2-5}
\cmidrule(lr){6-9}
\cmidrule(lr){10-10}
\cmidrule(lr){11-13}
\cmidrule(lr){14-14}
&
Live Para
& Live P-Mul
& Non-Live Para
& Non-Live P-Mul
& AIME24
& AIME25
& AMC23
& MATH500
& GPQA-D
& LiveCodeBench
& HumanEval+
& MBPP+
& Overall Avg. \\
\midrule

\rowcolor{groupgray}
\multicolumn{14}{c}{\textit{Base and Expert Models}} \\

\textbf{Qwen2.5-7B-Base}
& 56.25
& 29.17
& 63.00
& 34.00
& 6.67
& 0.00
& 32.50
& 54.80
& 30.13
& 8.22
& 70.73
& 35.19
& 34.32 \\

\textbf{ReTool-SFT}
& 81.25
& 58.33
& 83.00
& 79.00
& 10.00
& 13.33
& 47.50
& 55.20
& 9.09
& 12.92
& 31.10
& 44.18
& 36.35 \\

\textbf{AceInstruct}
& 12.50
& 29.17
& 50.00
& 52.00
& 6.67
& 6.67
& 45.00
& 72.40
& 18.69
& 29.35
& 66.46
& 70.11
& 35.65 \\

\textbf{ODA-Math}
& 25.00
& 8.33
& 20.00
& 10.00
& 46.67
& 33.33
& 85.00
& 69.60
& 8.25
& 7.83
& 35.37
& 39.95
& 27.61 \\

\midrule
\rowcolor{groupgray}
\multicolumn{14}{c}{\textit{Merged Models}} \\

\textbf{TA}
& \secondscore{43.75}
& \firstscore{54.17}
& \firstscore{79.50}
& \firstscore{76.50}
& \firstscore{20.00}
& \firstscore{16.67}
& \firstscore{70.00}
& \secondscore{75.00}
& \secondscore{29.63}
& \firstscore{20.94}
& \firstscore{74.39}
& \secondscore{63.76}
& \firstscore{47.89} \\

\textbf{TSV-Merge}
& \firstscore{50.00}
& \thirdscore{45.83}
& \thirdscore{75.00}
& \secondscore{72.50}
& \secondscore{16.67}
& \firstscore{16.67}
& \secondscore{65.00}
& \firstscore{75.20}
& \firstscore{30.64}
& \secondscore{19.96}
& \secondscore{73.78}
& \thirdscore{62.70}
& \secondscore{46.75} \\

\midrule
\textbf{\textsc{ResMerge} (Ours)}
& \secondscore{43.75}
& \secondscore{50.00}
& \secondscore{76.00}
& \thirdscore{71.00}
& \secondscore{16.67}
& \secondscore{13.33}
& \secondscore{65.00}
& \thirdscore{74.40}
& \thirdscore{28.11}
& \thirdscore{16.63}
& \thirdscore{72.56}
& \firstscore{65.34}
& \thirdscore{45.54} \\

\bottomrule
\end{tabular}
}
\caption{
Model merging results on the Qwen2.5-7B-Base SFT expert group across
individual benchmarks and the overall average. Shaded cells denote the
top distinct results among merged models in each column:
\colorbox{rankfirst}{\textbf{1st}},
\colorbox{ranksecond}{2nd}, and
\colorbox{rankthird}{3rd}; tied scores share the same color.
}
\label{tab:main_results_qwen25_base_sft}
\end{table*}

\paragraph{SFT-only expert group.}
Table~\ref{tab:main_results_qwen25_base_sft} reports results on three
SFT experts derived from Qwen2.5-7B-Base. \textsc{ResMerge} achieves the
best score on MBPP+ and the second-best score on six additional
benchmarks, but its Overall Avg.\ of 45.54 remains 1.21 and 2.35 points
below TSV-Merge and TA, respectively. Thus, while the method remains
competitive on several individual benchmarks, it does not retain the
consistent overall advantage observed for RL-post-trained experts. This
attenuation is consistent with the spectral differences between RL and
SFT task vectors discussed in Section~\ref{sec:spectral_structure_analysis},
suggesting that the head--residual structure exploited by
\textsc{ResMerge} is less pronounced in the SFT-only setting.

\subsection{Statistical Reliability and Expert-Capability Retention}
\label{sec:app_reliability_retention}

We complement the average benchmark results with two analyses. First, we
examine whether the gains of \textsc{ResMerge} remain stable under
changes in benchmark composition. Second, we evaluate whether these gains
are accompanied by balanced preservation of the specialized capabilities
of the source experts.

\paragraph{Statistical reliability.}
We conduct paired, domain-stratified benchmark bootstrap and
leave-one-benchmark-out (LOO) analyses against the strongest non-ours
method in each expert group. Following the main evaluation protocol,
benchmark-level differences are first averaged within each capability
domain and then equally averaged across domains. Detailed definitions are
provided in Appendix~\ref{app:geometry_and_metrics}.

As shown in Table~\ref{tab:statistical_reliability}, \textsc{ResMerge}
achieves positive full-suite gains across all three Qwen-based RL expert
groups. The corresponding $95\%$ bootstrap intervals lie entirely above
zero, and at least $99.76\%$ of the bootstrap replicates retain a
positive gain. The improvement also remains positive after removing any
single benchmark, with minimum LOO gains between $0.82$ and $1.67$
points. These results indicate that the observed improvements are stable
across benchmark resampling and are not driven by any individual
benchmark.

\begin{table*}[t]
\centering
\small
\renewcommand{\arraystretch}{1.12}
\setlength{\tabcolsep}{7pt}
\begin{tabular}{lccccc}
\toprule
\textbf{Expert Group}
& \textbf{Full $\Delta$}
& \textbf{95\% Bootstrap CI}
& \textbf{$P_{\mathrm{boot}}(\Delta>0)$}
& \textbf{Positive LOO}
& \textbf{Min LOO $\Delta$} \\
\midrule
Qwen2.5-7B-Base
& $+2.22$
& $[+1.02,\,+3.46]$
& $100.00\%$
& $12/12$
& $+1.67$ \\

Qwen2.5-7B-Instruct
& $+1.25$
& $[+0.35,\,+2.12]$
& $99.76\%$
& $12/12$
& $+0.82$ \\

Qwen3-4B-Base
& $+1.77$
& $[+0.89,\,+2.70]$
& $100.00\%$
& $12/12$
& $+1.44$ \\
\bottomrule
\end{tabular}
\caption{
Benchmark-composition reliability of \textsc{ResMerge} relative to the
strongest non-ours method on three Qwen-based RL expert groups, based on
$10{,}000$ paired, domain-stratified bootstrap replicates and LOO
analysis.
}
\label{tab:statistical_reliability}
\end{table*}

\paragraph{Expert-capability retention.}
Overall averages can conceal capability trade-offs, since stronger
aggregate performance may coincide with substantial degradation on the
specialized domain of an individual expert. We therefore compare
\textsc{ResMerge} with the strongest non-ours method in each group using
Mean ECR, Min ECR, and Worst Capability Shortfall (WCS).

Table~\ref{tab:expert_capability_retention} shows that \textsc{ResMerge} consistently improves all three metrics across the three Qwen-based RL expert groups. Compared with the corresponding non-ours method, the improvement in Mean ECR ranges from 3.1 to 5.9 percentage points, while the improvement in Min ECR ranges from 4.3 to 8.4 percentage points. WCS is also reduced by 1.51 to 3.05 points across the three settings. The largest improvement in worst-case relative retention is observed on Qwen2.5-7B-Instruct, where Min ECR increases from $0.832$ to $0.916$. These results indicate that the overall gains of \textsc{ResMerge} are accompanied by more balanced preservation of source-expert capabilities.

\begin{table*}[t]
\centering
\small
\renewcommand{\arraystretch}{1.12}
\setlength{\tabcolsep}{9pt}
\begin{tabular}{llccc}
\toprule
\textbf{Expert Group}
& \textbf{Method}
& \textbf{Mean ECR $\uparrow$}
& \textbf{Min ECR $\uparrow$}
& \textbf{WCS $\downarrow$} \\
\midrule

Qwen2.5-7B-Base
& Best Non-Ours (ISO-C)
& 0.923
& 0.862
& 4.88 \\
&
\textbf{\textsc{ResMerge} (Ours)}
& \textbf{0.982}
& \textbf{0.918}
& \textbf{3.37} \\

\addlinespace[2pt]
Qwen2.5-7B-Instruct
& Best Non-Ours (TIES)
& 0.948
& 0.832
& 5.05 \\
&
\textbf{\textsc{ResMerge} (Ours)}
& \textbf{0.979}
& \textbf{0.916}
& \textbf{2.52} \\

\addlinespace[2pt]
Qwen3-4B-Base
& Best Non-Ours (TSV-Merge)
& 0.886
& 0.774
& 16.19 \\
&
\textbf{\textsc{ResMerge} (Ours)}
& \textbf{0.920}
& \textbf{0.817}
& \textbf{13.14} \\

\bottomrule
\end{tabular}
\caption{
Expert-capability retention across three Qwen-based RL expert groups.
Higher ECR and lower WCS indicate better capability preservation; bold
marks the better result within each group.
}
\label{tab:expert_capability_retention}
\end{table*}

Together, these analyses show that the gains of \textsc{ResMerge} are
robust to benchmark composition and do not come at the expense of
severely degrading individual source-expert capabilities.

\end{document}